\documentclass{article}

\usepackage{arxiv}

\usepackage[utf8]{inputenc}
\usepackage[T1]{fontenc}

\usepackage{url}
\usepackage{microtype}
\usepackage{textcomp}
\usepackage{xspace}

\usepackage{amsmath}
\usepackage{amssymb}
\usepackage{amsfonts}

\usepackage{booktabs}
\usepackage{multirow}
\usepackage{siunitx}
\usepackage{adjustbox}

\usepackage{graphicx}
\usepackage{subcaption}
\usepackage{wrapfig}
\usepackage{pdflscape}

\graphicspath{{./figures/}{./images/}}

\usepackage{algorithm}
\usepackage{algpseudocode}

\usepackage[dvipsnames,table,x11names]{xcolor}
\usepackage{tikz}
\usepackage{pgfplots}
\usepackage{pgfplotstable}

\usetikzlibrary{pgfplots.groupplots,fillbetween}
\usepgfplotslibrary{groupplots}
\usepgfplotslibrary{fillbetween}
\usepgfplotslibrary{polar}

\pgfplotsset{compat=1.18}

\usepackage{hyperref}

\definecolor{arylideyellow}{rgb}{0.91, 0.84, 0.42}

\definecolor{pastelgreen}{RGB}{180, 235, 200}
\definecolor{pastelyellow}{RGB}{255, 245, 190}
\definecolor{pastelorange}{RGB}{255, 210, 160}
\definecolor{pastelred}{RGB}{255, 170, 150}

\definecolor{lemonade}{HTML}{FBD271}
\definecolor{seaBlue}{HTML}{0F4E77}
\definecolor{matcha}{HTML}{89A377}
\definecolor{honeyDrizzle}{HTML}{E7BD8B}
\definecolor{coffeeGrounds}{HTML}{372516}
\definecolor{oatMilk}{HTML}{EFE4C8}
\definecolor{berryClay}{HTML}{9B5A5A}
\definecolor{ultraViolet}{HTML}{595082}
\definecolor{darkPurple}{HTML}{2C263F}
\definecolor{LlamaProp}{RGB}{86,99,182}

\newcommand{\heat}[1]{%
    \ifdim #1 pt > 73pt \cellcolor{pastelgreen}{#1}%
    \else\ifdim #1 pt > 70pt \cellcolor{pastelyellow}{#1}%
    \else\ifdim #1 pt > 67pt \cellcolor{pastelorange}{#1}%
    \else \cellcolor{pastelred}{#1}%
    \fi\fi\fi
}

\pgfplotscreateplotcyclelist{color list}{
  {blue},
  {red},
  {green!50!black},
  {orange},
  {purple},
  {cyan!60!black},
  {magenta},
  {brown}
}

\newcommand{\opac}{0.15}
\newcommand{\stempcol}{LightGoldenrod2}
\newcommand{\dstempcol}{DarkGoldenrod2}
\newcommand{\deepseekcol}{DarkOliveGreen3}
\newcommand{\llamacol}{SteelBlue4}
\newcommand{\qwencol}{DarkOrchid3}
\newcommand{\mixtralcol}{CadetBlue3}
\newcommand{\llavacol}{IndianRed2}
\newcommand{\proposedcol}{LlamaProp}
\newcommand{\lw}{0.5pt}

\newlength\colgap
\newlength\rowgap
\newcommand{\SC}[1]{\textsc{#1}}

\newcommand{\dst}{\texttt{DST}\xspace}
\newcommand{\st}{\texttt{ST}\xspace}
\newcommand{\llama}{\texttt{LLaMA-3.1}\xspace}
\newcommand{\deepseek}{\texttt{DeepSeek}\xspace}
\newcommand{\qwen}{\texttt{Qwen-2.5}\xspace}
\newcommand{\mixtral}{\texttt{Mixtral}\xspace}
\newcommand{\rsllava}{\texttt{RS-LLaVA}\xspace}

\newcommand{\rsscn}{\texttt{RSSCN7}\xspace}
\newcommand{\eurosat}{\texttt{EuroSAT}\xspace}
\newcommand{\rsc}{\texttt{RSC11}\xspace}
\newcommand{\siriwhu}{\texttt{SIRI-WHU}\xspace}
\newcommand{\whurs}{\texttt{WHU-RS19}\xspace}
\newcommand{\aid}{\texttt{AID}\xspace}
\newcommand{\optimal}{\texttt{Optimal-31}\xspace}
\newcommand{\rsicblow}{\texttt{RSICB128}\xspace}
\newcommand{\rsicbhigh}{\texttt{RSICB256}\xspace}
\newcommand{\patternnet}{\texttt{PatternNet}\xspace}
\newcommand{\resisc}{\texttt{RESISC45}\xspace}
\newcommand{\mlrsnet}{\texttt{MLRSNet}\xspace}

\newlength{\retrievallabeltextwidth}
\newlength{\retrievallabelwidth}
\newlength{\retrievallabelheight}
\newlength{\retrievallabelfontsize}
\newlength{\retrievallabellineskip}

\newcommand{\imgneglabel}[2]{%
    \begin{tikzpicture}[baseline=0pt]
        \node[inner sep=0pt, anchor=south] at (0,0) {\imgneg{#1}};

        \node[
            overlay,
            anchor=south,
            fill=black!70,
            fill opacity=0.7,
            text=white,
            text opacity=1,
            font=\fontsize{\the\retrievallabelfontsize}{\the\retrievallabellineskip}\selectfont,
            align=center,
            inner sep=0pt,
            outer sep=0pt,
            minimum width=\retrievallabelwidth,
            minimum height=\retrievallabelheight,
            rounded corners=1pt
        ] at (0,4pt) {%
            \adjustbox{max width=\retrievallabeltextwidth}{#2}%
        };
    \end{tikzpicture}%
}

\title{Beyond Templates: Revisiting Zero-Shot Remote Sensing through Meta-Prompting}

\author{
Eirini Baltzi \\
Remote Sensing Lab, \\
National Technical University of Athens \\
Athens, Greece
\And
Dionysis Christopoulos \\
Remote Sensing Lab, \\
National Technical University of Athens \\
Athens, Greece
\And
Sotiris Spanos \\
Remote Sensing Lab, \\
National Technical University of Athens \\
Athens, Greece
\And
Valsamis Ntouskos \\
Department of Engineering and Sciences, \\
Universitas Mercatorum \\
Rome, Italy
\And
Konstantinos Karantzalos \\
Remote Sensing Lab, \\
National Technical University of Athens \\
Athens, Greece
}

\begin{document}

\maketitle

% --------------------------------------------------
% Abstract
% --------------------------------------------------
\begin{abstract}
Vision-language models (VLMs) have sparked growing interest in zero-shot Earth Observation (EO) downstream tasks, with further gains enabled by remote-sensing-adapted models. We examine this setting across 17 VLM variants and 12 remote sensing (RS) datasets under Meta-Prompting for Visual Recognition (MPVR), and show that zero-shot performance remains highly sensitive to textual design choices, from the meta-prompts used to guide the LLM in generating class descriptions to the descriptions themselves. We explore why semantically rich LLM-generated class descriptions do not translate into consistent gains over simple domain-adapted CLIP-style descriptions. While LLM descriptions are more semantically expressive, they can also introduce noise in the text embedding space, reducing robustness in downstream tasks. We support this observation through a text log-likelihood analysis in the whitened CLIP feature space, comparing LLM-generated and template-based descriptions. Building on this finding, we study query embedding calibration and show that lightweight calibration of the query space consistently yields strong improvements in zero-shot classification and retrieval. Overall, our results provide practical insight into the trade-off between semantic richness and robustness, and identify embedding calibration as a simple and effective tool for improving zero-shot remote sensing performance.
\end{abstract}


% --------------------------------------------------
% Eye-catcher figure
% --------------------------------------------------
\begin{figure}[ht!]
    \centering
    \begin{tikzpicture}
\begin{polaraxis}[
    width=8cm, 
    height=8cm,
    ymin=0, ymax=0.18,
    ytick={},
    yticklabel=\empty, 
    grid=none,
    axis background/.style={},
    axis line style={draw=none},
    y filter/.expression={y - 0.57},
    xtick={0,22.5,45,67.5,90,112.5,135,157.5,180,202.5,225,247.5,270,292.5,315,337.5},
    % xticklabels={
    %   CLIP-B16,CLIP-L14,MetaCLIP400M-B16,MetaCLIP400M-L14,MetaCLIP2.5B-B16,MetaCLIP2.5B-L14,SigLIP-B16,SigLIP-L16,SigLIP2-B16,SigLIP2-L16,RemoteCLIP-B32,RemoteCLIP-L14,GeoRSCLIP-B32,GeoRSCLIP-L14,LAIONCLIP-L14,SkyCLIP-L14
    % },
    xticklabels=\empty,
    xticklabel style={font=\scriptsize},
    enlargelimits=false,
    clip=false, % allow labels to extend beyond the circle
    legend style={
        font=\scriptsize,
        draw=none,
        fill=none,
        at={(0.5,-0.12)},
        anchor=north,
        legend columns=4,
        /tikz/every even column/.append style={column sep=6pt}
    },
    legend image code/.code={
        \draw[#1, fill opacity=\opac, line width=\lw]
            (-0.20cm,-0.15cm) rectangle (0.3cm,0.15cm);
    },
]

% ---- Outer ring as discrete polygon (one vertex per model)
\addplot[
    thick,
    color=black,
    fill=orange!3,
    mark=none,
    forget plot,
]
coordinates {
 (0,0.75)
 (22.5,0.75)
 (45,0.75)
 (67.5,0.75)
 (90,0.75)
 (112.5,0.75)
 (135,0.75)
 (157.5,0.75)
 (180,0.75)
 (202.5,0.75)
 (225,0.75)
 (247.5,0.75)
 (270,0.75)
 (292.5,0.75)
 (315,0.75)
 (337.5,0.75)
} -- cycle;

% =====================================================
% Polygonal radial grids (discrete rings)
% =====================================================
\pgfplotsinvokeforeach{0.600,0.625,0.650,0.675,0.700,0.725}{
    \addplot[
        draw=gray!15,
        line width=0.3pt,
        mark=none,
        forget plot,
    ]
    coordinates {
        (0,#1)
        (22.5,#1)
        (45,#1)
        (67.5,#1)
        (90,#1)
        (112.5,#1)
        (135,#1)
        (157.5,#1)
        (180,#1)
        (202.5,#1)
        (225,#1)
        (247.5,#1)
        (270,#1)
        (292.5,#1)
        (315,#1)
        (337.5,#1)
    } -- cycle;
}

% =====================================================
% Radial spokes (center to each xtick)
% =====================================================
\pgfplotsinvokeforeach{0,22.5,45,67.5,90,112.5,135,157.5,180,202.5,225,247.5,270,292.5,315,337.5}{
    \addplot[
        draw=gray!15,
        line width=0.3pt,
        mark=none,
        forget plot,
    ]
    coordinates {
        (#1,0.57)
        (#1,0.75)
    };
}

% ---- ds-temp
\addplot+[mark=none,solid, draw=\dstempcol,
    fill=\dstempcol,
    fill opacity=\opac,
    line width=\lw]
coordinates {
 (0,0.5994)
 (22.5,0.6640)
 (45,0.6313)
 (67.5,0.6532)
 (90,0.6413)
 (112.5,0.6855)
 (135,0.5953)
 (157.5,0.6308)
 (180,0.6396)
 (202.5,0.6829)
 (225,0.6124)
 (247.5,0.6486)
 (270,0.6506)
 (292.5,0.7131)
 (315,0.6894)
 (337.5,0.7073)
} -- cycle;
\addlegendentry{\dst}

% ---- deepseek_7b
\addplot+[mark=none,solid,draw=\deepseekcol,
    fill=\deepseekcol,
    fill opacity=\opac,
    line width=\lw]
coordinates {
 (0,0.5967)
 (22.5,0.6426)
 (45,0.6259)
 (67.5,0.6378)
 (90,0.6173)
 (112.5,0.6660)
 (135,0.5879)
 (157.5,0.6452)
 (180,0.6219)
 (202.5,0.6671)
 (225,0.6382)
 (247.5,0.6463)
 (270,0.6682)
 (292.5,0.7086)
 (315,0.6701)
 (337.5,0.7063)
} -- cycle;
\addlegendentry{\deepseek}

% ---- llama3.1
\addplot+[mark=none,solid, draw=\llamacol,
    fill=\llamacol,
    fill opacity=\opac,
    line width=\lw]
coordinates {
 (0,0.6031)
 (22.5,0.6369)
 (45,0.6373)
 (67.5,0.6540)
 (90,0.6350)
 (112.5,0.6752)
 (135,0.5966)
 (157.5,0.6574)
 (180,0.6294)
 (202.5,0.6793)
 (225,0.6355)
 (247.5,0.6570)
 (270,0.6714)
 (292.5,0.7098)
 (315,0.6716)
 (337.5,0.7011)
} -- cycle;
\addlegendentry{\llama}

% ---- qwen2.5
\addplot+[mark=none,solid,draw=\qwencol,
    fill=\qwencol,
    fill opacity=\opac,
    line width=\lw]
coordinates {
 (0,0.5927)
 (22.5,0.6454)
 (45,0.6236)
 (67.5,0.6457)
 (90,0.6280)
 (112.5,0.6639)
 (135,0.5821)
 (157.5,0.6501)
 (180,0.6248)
 (202.5,0.6693)
 (225,0.6272)
 (247.5,0.6567)
 (270,0.6590)
 (292.5,0.7028)
 (315,0.6766)
 (337.5,0.7081)
} -- cycle;
\addlegendentry{\qwen}

% ---- mixtral
\addplot+[mark=none,solid,draw=\mixtralcol,
    fill=\mixtralcol,
    fill opacity=\opac,
    line width=\lw]
coordinates {
 (0,0.5927)
 (22.5,0.6535)
 (45,0.6100)
 (67.5,0.6443)
 (90,0.6121)
 (112.5,0.6554)
 (135,0.5727)
 (157.5,0.6232)
 (180,0.6061)
 (202.5,0.6623)
 (225,0.6478)
 (247.5,0.6501)
 (270,0.6640)
 (292.5,0.7159)
 (315,0.6695)
 (337.5,0.7077)
} -- cycle;
\addlegendentry{\mixtral}

% ---- rsllava
\addplot+[mark=none,solid,draw=\llavacol,
    fill=\llavacol,
    fill opacity=\opac,
    line width=\lw]
coordinates {
 (0,0.5845)
 (22.5,0.6410)
 (45,0.6158)
 (67.5,0.6337)
 (90,0.6281)
 (112.5,0.6630)
 (135,0.5925)
 (157.5,0.6459)
 (180,0.6280)
 (202.5,0.6748)
 (225,0.6372)
 (247.5,0.6415)
 (270,0.6632)
 (292.5,0.7148)
 (315,0.6710)
 (337.5,0.7070)
} -- cycle;
\addlegendentry{\rsllava}

% ---- llama3.1 -- Centered
\addplot+[mark=none,solid, draw=\proposedcol,
    fill=\proposedcol,
    fill opacity=\opac,
    line width=\lw]
coordinates {
 (0,0.6368)
 (22.5,0.6845)
 (45,0.6631)
 (67.5,0.6827)
 (90,0.6647)
 (112.5,0.7133)
 (135,0.6183)
 (157.5,0.6792)
 (180,0.6532)
 (202.5,0.7125)
 (225,0.6482)
 (247.5,0.6876)
 (270,0.6902)
 (292.5,0.7412)
 (315,0.7015)
 (337.5,0.7105)
} -- cycle;
\addlegendentry{\llama \texttt{calibrated}}

% =====================================================
% Manual xtick labels (uniform radius, easy tweaking)
% =====================================================
\node[font=\scriptsize, anchor=west,  xshift=2pt, yshift=0pt] at (axis cs:0,0.18)     {CLIP$_{\text{ViT/B}}$};
\node[font=\scriptsize, anchor=west,  xshift=2pt, yshift=4pt] at (axis cs:22.5,0.18)  {CLIP$_{\text{ViT/L}}$};

\node[font=\scriptsize, anchor=south west, xshift=2pt, yshift=-4pt] at (axis cs:45,0.18)   {MetaCLIP-400M$_{\text{ViT/B}}$};
\node[font=\scriptsize, anchor=south west, xshift=0pt, yshift=0pt] at (axis cs:67.5,0.18) {MetaCLIP-400M$_{\text{ViT/L}}$};

\node[font=\scriptsize, anchor=south, xshift=0pt, yshift=4pt] at (axis cs:90,0.18)   {MetaCLIP-2.5B$_{\text{ViT/B}}$};
\node[font=\scriptsize, anchor=south east, xshift=0pt, yshift=0pt] at (axis cs:112.5,0.18){MetaCLIP-2.5B$_{\text{ViT/L}}$};

\node[font=\scriptsize, anchor=south east,  xshift=-2pt, yshift=-4pt] at (axis cs:135,0.18)  {SigLIP$_{\text{ViT/B}}$};
\node[font=\scriptsize, anchor=east,  xshift=-2pt, yshift=4pt] at (axis cs:157.5,0.18){SigLIP$_{\text{ViT/L}}$};

\node[font=\scriptsize, anchor=east,  xshift=-2pt, yshift=0pt] at (axis cs:180,0.18)  {SigLIP2$_{\text{ViT/B}}$};
\node[font=\scriptsize, anchor=east,  xshift=-2pt, yshift=-2pt] at (axis cs:202.5,0.18){SigLIP2$_{\text{ViT/L}}$};

\node[font=\scriptsize, anchor=north east, xshift=-2pt, yshift=4pt] at (axis cs:225,0.18)  {RemoteCLIP$_{\text{ViT/B}}$};
\node[font=\scriptsize, anchor=north east, xshift=0pt, yshift=0pt] at (axis cs:247.5,0.18){RemoteCLIP$_{\text{ViT/L}}$};

\node[font=\scriptsize, anchor=north, xshift=0pt, yshift=-4pt] at (axis cs:270,0.18)  {GeoRSCLIP$_{\text{ViT/B}}$};
\node[font=\scriptsize, anchor=north west, xshift=0pt, yshift=0pt] at (axis cs:292.5,0.18){GeoRSCLIP$_{\text{ViT/L}}$};

\node[font=\scriptsize, anchor=north west,  xshift=2pt, yshift=4pt] at (axis cs:315,0.18)  {CLIP LAION RS$_{\text{ViT/L}}$};
\node[font=\scriptsize, anchor=west,  xshift=2pt, yshift=-2pt] at (axis cs:337.5,0.18){SkyCLIP$_{\text{ViT/L}}$};

\end{polaraxis}
\end{tikzpicture}
    \vspace{-2pt}
    \caption{Comparison of template-based \dst and meta-prompting approaches in terms of classification accuracy across the evaluated VLMs. We also report the \llama variant obtained by embedding calibration. \textcolor{\proposedcol}{\llama \texttt{calibrated}} consistently defines the strongest overall performance regime, surpassing all other prompt sources.}
    \label{fig:placeholder}
    \vspace{-10pt}
\end{figure}
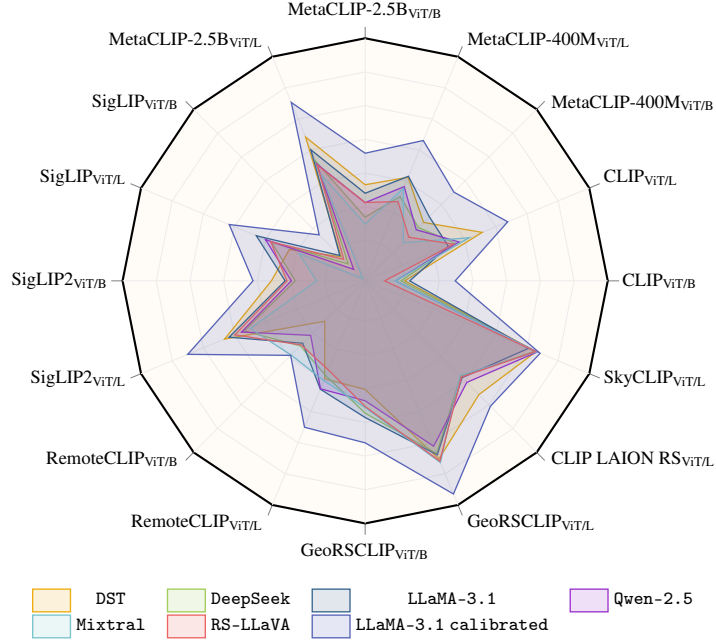

% --------------------------------------------------
% Main paper
% --------------------------------------------------
\section{Introduction}
\label{sec:intro}

Vision-language models (VLMs), such as CLIP~\cite{clip}, enable zero-shot visual recognition by aligning image and text representations in a shared embedding space. In this setting, classification is performed by comparing image embeddings with textual descriptions of candidate classes. While this paradigm has demonstrated strong generalization on natural-image benchmarks, its behavior in remote sensing (RS) imagery remains underexplored and highly sensitive to prompt formulation. Although RS datasets are often included in broader VLM evaluations, our work provides a targeted, domain-specific analysis of these dynamics.

Remote sensing scenes differ substantially from natural images due to overhead viewing geometry, varying spatial resolution, and scene-level semantics. As a result, the generic templates commonly used in zero-shot evaluation (e.g., ``a photo of a {class}'') often fail to capture RS-specific visual cues. Recent work on Meta-Prompting for Visual Recognition (MPVR)~\cite{mirza2024mpvr} has shown that large language models (LLMs) can automatically generate class- and dataset-aware descriptions, enriching the textual supervision used for zero-shot prediction. Our prior work \cite{metaprompting2026} establishes the feasibility of using MPVR-style prompting with open-source LLMs to generate remote-sensing-aware class descriptions. However, we observed that performance gains vary across datasets and vision-language backbones.

% However, the behavior of MPVR-style prompting under the domain shift induced by aerial and satellite imagery remains insufficiently understood. In particular, it is still unclear how generated descriptions interact with prompt aggregation, embedding-space geometry, and downstream task formulation in remote sensing.

% In this work, we study meta-prompting for zero-shot remote sensing through a comprehensive empirical evaluation across twelve RS datasets and a broad range of vision-language models. Building on the two-stage prompt generation procedure introduced in MPVR, we adapt prompt construction to remote-sensing scene semantics and examine several components that affect zero-shot performance. Specifically, we investigate the role of prompt aggregation, including a top-$k$ voting variant in addition to standard prototype averaging, and we study embedding calibration strategies based on image- and text-side centering, analyzing their asymmetric impact on classification and text-to-image retrieval.

In this work, we revisit meta-prompting for zero-shot remote sensing through a comprehensive empirical evaluation across twelve RS datasets and a broad range of vision-language models. Building on the two-stage prompt generation procedure introduced in MPVR, we adapt prompt construction to remote-sensing scene semantics and examine several components that affect zero-shot performance. Specifically, we investigate the role of prompt aggregation, including a top-$k$ voting variant in addition to prototype averaging, and we study embedding calibration strategies based on image- and text-side centering, analyzing their asymmetric impact on classification and text-to-image retrieval.

Our study provides practical insights into how textual diversity, aggregation, and representation geometry interact in zero-shot remote sensing. The results indicate that prompt design and lightweight calibration substantially influence both performance and robustness across datasets and model families, offering useful guidance for deploying VLMs in RS applications and for future work on prompt-based adaptation under domain shift.

The main contributions of this work are:
\begin{itemize}
    \item \textbf{Large-scale RS study of meta-prompting.} We evaluate MPVR-style prompting across 12 remote-sensing datasets and 17 vision-language model variants.
    \item \textbf{Embedding calibration analysis.} We study simple embedding-centering strategies and show task-dependent effects: image centering consistently benefits classification, while text centering is more effective for retrieval.
    \item \textbf{Dual-task evaluation.} Beyond zero-shot classification, we report text-to-image retrieval results, providing a complementary lens for assessing prompt quality and cross-modal behavior in RS.
    \item \textbf{Analysis of prompt robustness and embedding geometry.} We study why semantically rich LLM-generated descriptions can be less stable than templates by analyzing their behavior in the whitened CLIP feature space.
\end{itemize}

\section{Related Work}

\paragraph{Vision-Language Models for Zero-Shot Recognition}
Vision-language models (VLMs), such as CLIP~\cite{clip}, MetaCLIP~\cite{xu2023demystifying}, and SigLIP~\cite{siglip}, have established a powerful paradigm for zero-shot visual recognition by learning a shared embedding space between images and natural language. In this setting, classification is performed by matching the image representations with the textual descriptions of candidate classes, enabling strong generalization to unseen categories~\cite{clip}. Despite their success on natural image benchmarks, VLMs remain highly sensitive to the formulation of textual prompts~\cite{zhou2022learning,zhou2022conditional}, and their performance can degrade under domain shift, especially when the target visual distribution differs substantially from that observed during pretraining~\cite{wortsman2022robust}. While much of the literature has focused on improving model architectures or scaling pretraining data, comparatively less attention has been given to understanding how the prompt design affects zero-shot behavior in domain-specific settings.

\paragraph{Vision-Language Models in Remote Sensing}
Applying VLMs to remote sensing (RS) imagery introduces additional challenges due to the unique properties of overhead data, including varying spatial resolutions, scale ambiguity, and scene-level semantics~\cite{li2018deep}. Recent efforts such as RemoteCLIP~\cite{liu2023remoteclip}, GeoRSCLIP~\cite{zhang2024rs5m}, and SkyCLIP~\cite{wang2024skyscript} adapt vision-language models to RS data through domain-specific pretraining or large-scale remote sensing vision-language corpora. These models consistently improve zero-shot performance relative to general-purpose backbones, yet the mismatch between generic textual supervision and RS-specific visual content remains an important limitation. This motivates not only the use of domain-aware prompt design, but also a broader analysis of how prompting strategies interact with model domain adaptation in remote sensing.

\paragraph{Prompt Engineering and Meta-Prompting}
Prompt engineering has emerged as a central component of zero-shot evaluation, with prior work exploring manually designed templates, prompt ensembling, and automatic prompt generation~\cite{clip,zhou2022learning,zhou2022conditional}. Among these directions, meta-prompting approaches such as MPVR~\cite{mirza2024mpvr} are particularly relevant, as they use large language models (LLMs) to generate richer and more diverse class descriptions than fixed templates. By increasing textual coverage, such methods can improve alignment between text and image embeddings, especially when simple prompts fail to capture the relevant semantics of a class. However, the benefit of richer descriptions is not uniform, and their interaction with the underlying embedding space remains poorly understood, particularly under the domain shift induced by aerial and satellite imagery.

\paragraph{Large Language Models as Prompt Generators in Remote Sensing}
Recent advances in large language models have enabled the automatic generation of detailed textual descriptions that can serve to guide frozen VLMs in zero-shot settings. These descriptions may incorporate attributes, contextual cues, and domain-relevant scene characteristics that go beyond class-name-based templates~\cite{mirza2024mpvr}. In remote sensing, this is especially appealing, as overhead scenes often require descriptions that capture spatial layout, land-cover composition, and imaging perspective~\cite{li2018deep,lu2017exploring}. In this context, recent open-source and domain-specialized LLMs, including LLaMA-3.1~\cite{llama3_1}, DeepSeek~\cite{bi2024deepseek}, Qwen-2.5~\cite{qwen2_5}, Mixtral~\cite{mixtral}, and RS-LLaVA~\cite{bazi2024rs}, provide a broad space of prompt-generation sources to study. At the same time, the use of LLM-generated prompts in RS remains underexplored from a comparative and analytical perspective. While individual aspects have been explored, a systematic understanding of how different prompt sources interact with both general-purpose and remote-sensing-adapted VLMs remains limited, particularly regarding how the geometry of the resulting text representations affects downstream behavior across tasks. In this work, we address this gap through a large-scale evaluation of template-based and meta-prompt-generated descriptions across multiple RS datasets, VLM families, open-source LLMs, and downstream tasks, while also examining the role of embedding calibration in both classification and retrieval.

\section{Evaluation Framework}
\label{sec:method}

\begin{figure*}[ht!] 
    \centering
    \includegraphics[width=\textwidth]{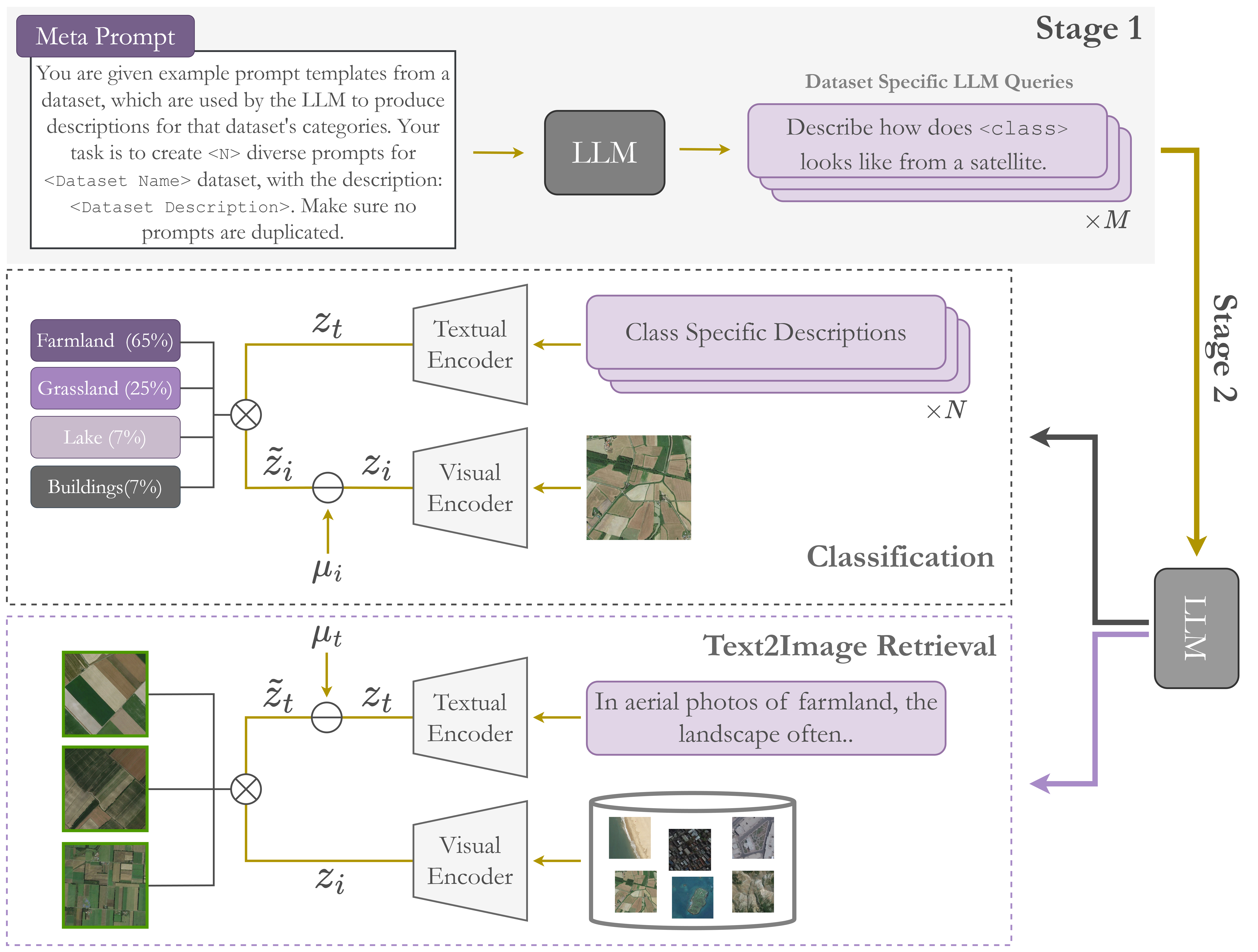} 
    \caption{Overview of our evaluation framework. In Stage~1, dataset-level information is used to generate dataset-specific LLM queries. In Stage~2, these queries produce class-specific descriptions, which are encoded by the frozen text encoder, while images are mapped into the same shared embedding space by the frozen visual encoder. For zero-shot classification, image embeddings are centered before matching with aggregated text prototypes. For text-to-image retrieval, text embeddings are centered before matching with image embeddings. Here, $\mu_i$ and $\mu_t$ denote the mean embedding vectors computed over the query modality of the corresponding downstream task, namely the set of image embeddings for classification and the set of text embeddings for retrieval.}
    \label{fig:eval_fr}
\end{figure*}

To systematically study the impact of textual supervision on zero-shot performance in remote sensing, we build upon the unified evaluation framework introduced in MPVR~\cite{mirza2024mpvr}. We formulate our zero-shot remote sensing evaluation pipeline as two explicit stages (Figure ~\ref{fig:eval_fr}). In the first stage, we contextualize a selected LLM using dataset-level information to generate dataset-specific queries. In our setup, GPT-5 is used for this meta-prompting step. Subsequently, these queries are used to prompt open-source LLMs, which produce class-specific textual descriptions for the categories of each downstream target dataset. In the second stage, the generated descriptions are encoded with the frozen VLM text encoder and compared against image embeddings extracted from the frozen VLM visual encoder. Unless stated otherwise, we apply prompt embedding aggregation at the class level and a simple yet highly effective calibration step in the query modality before computing zero-shot predictions. This formulation provides a controlled framework for analyzing how dataset-level contextualization, class-description generation, prompt aggregation, and query-side embedding calibration affect downstream zero-shot performance across different vision-language models (VLMs) and remote sensing benchmarks.

\paragraph{Downstream tasks}
We evaluate the generated class representations on two downstream tasks, namely zero-shot image classification and text-to-image retrieval. In the former, image embeddings are compared against the set of aggregated text prototypes, and classification is performed by identifying the class that maximizes similarity. In the latter, each class prototype acts as a text query to retrieve relevant images from the gallery by ranking image embeddings according to cosine similarity. Together, these tasks offer complementary perspectives on the quality of alignment in the shared vision-language embedding space.

\subsection{Description Generation Strategies}

\paragraph{Template-based descriptions}
As a point of comparison, we consider two forms of fixed textual supervision that bypass the dataset-specific query generation stage and construct descriptions directly from class names. In particular, the Simple Template (\st) baseline, uses a single generic template (``a satellite image of a \{class\}''), whereas the Domain-Specific Template (\dst) baseline, employs multiple remote-sensing aware templates designed to better reflect the overhead imaging perspective (e.g., ``an overhead image of a \{class\}''). These baselines provide simple, controlled, and interpretable textual descriptions, but remain limited to shallow class-name conditioning and therefore have reduced capacity to capture richer class semantics and intra-class visual variability.

\paragraph{LLM-based description generation}
Under the LLM-based description generation setting, we adopt a meta-prompting strategy (\cite{mirza2024mpvr}) in which dataset-level information is first used to formulate dataset-specific queries, and these queries are subsequently supplied to open-source LLMs to generate class-specific textual descriptions across the target label space. This way, the language model is not prompted directly with isolated class names, but is first contextualized at the dataset level, allowing the generated descriptions to better reflect the semantics and visual characteristics of the target benchmark. Compared with fixed template baselines, this meta-prompting strategy yields multiple descriptions per class with greater diversity in wording, semantic emphasis, and descriptive detail. In this study, we evaluate several open-source LLMs, including \llama~\cite{llama3_1}, \deepseek~\cite{bi2024deepseek}, \qwen~\cite{qwen2_5}, \mixtral~\cite{mixtral}, and the domain-specific \rsllava~\cite{bazi2024rs}. This design allows us to analyze how the source and characteristics of the meta-prompt-generated descriptions influence the resulting text embedding space and, in turn, zero-shot downstream performance. 

\subsection{Prototype construction}
\paragraph{Prompt Aggregation} Given a set of textual descriptions for each class, obtained using either the template- or LLM-based description generation strategy, we map them into the shared embedding space using the frozen text encoder of the corresponding vision-language model. Formally, let $\mathcal{T}_c = \{t_{c,1}, \dots, t_{c,N_c}\}$ denote the set of textual descriptions associated with class $c$, where $N_c$ is the number of descriptions for that class. Each description is encoded as
\[
z^{\text{text}}_{c,n} = f_{\text{text}}(t_{c,n}),
\]
where $f_{\text{text}}(\cdot)$ is the frozen text encoder. A single class representation is then formed by averaging the corresponding text embeddings,
\[
\bar{z}^{\text{text}}_c = \frac{1}{N_c}\sum_{n=1}^{N_c} z^{\text{text}}_{c,n},
\]
yielding one class prototype per category.

\paragraph{Embedding calibration} Following standard practice in vision-language models, both visual and textual embeddings are $\ell_2$-normalized, and cross-modal matching is performed using cosine similarity. Given an image $x$ with embedding
\[
z^{\text{img}} = f_{\text{img}}(x),
\]
the similarity between image $x$ and class $c$ is computed as
\[
s(x,c) = \cos\!\bigl(z^{\text{img}}, \bar{z}^{\text{text}}_c\bigr),
\]
where $f_{\text{img}}(\cdot)$ denotes the frozen visual encoder. Additionally, we apply a lightweight embedding calibration step based on mean-centering. Specifically, prior to similarity computation, we subtract the mean vector of the query modality, where the query side is defined by the downstream task. Let $\mathcal{Q} = \{q_1,\dots,q_M\}$ denote the set of query embeddings for the task under consideration. We compute the mean query vector as
\[
\mu_{\mathcal{Q}} = \frac{1}{M}\sum_{j=1}^{M} q_j,
\]
and center each query embedding according to
\[
\tilde{q}_j = q_j - \mu_{\mathcal{Q}}.
\]
For zero-shot image classification, this corresponds to centering image embeddings, whereas for text-to-image retrieval, it corresponds to centering text embeddings. In contrast, centering the database-side modality did not provide consistent gains and often reduced performance. We therefore apply calibration only to the query modality. This operation effectively suppresses dominant shared components that reflect generic semantics or modality-level bias rather than class-discriminative content, thereby improving the sensitivity of cosine similarity to meaningful cross-modal differences \cite{pmlr-v267-levi25b}.
\section{Experimental Setup}
\label{sec:setup}

\paragraph{Evaluated datasets} 
We conduct our evaluation on the full suite of remote sensing scene classification benchmarks adopted in RemoteCLIP~\cite{liu2023remoteclip}. This benchmark collection comprises 12 datasets with substantial variation in dataset scale, number of classes, image resolution, and ground sampling distance (GSD), thereby covering diverse scene categories and levels of semantic granularity. A detailed dataset overview is provided in the appendix (Table~\ref{tab:datasets_overview}). All metrics reported in the following section are obtained by evaluating on the entirety of each dataset, with every available image treated as part of the test set, since the proposed setting does not involve any task-specific training.

\paragraph{Evaluated Vision-Language Models} We evaluate a diverse set of vision-language models (VLMs) spanning both general-purpose and remote sensing-adapted architectures. The evaluated model families include CLIP~\cite{clip}, MetaCLIP~\cite{xu2023demystifying}, SigLIP~\cite{siglip}, and SigLIP2~\cite{siglip2}, as well as remote sensing-oriented models such as RemoteCLIP~\cite{liu2023remoteclip}, GeoRSCLIP~\cite{zhang2024rs5m}, CLIP-LAION-RS~\cite{wang2024skyscript}, and SkyCLIP~\cite{wang2024skyscript}. Whenever available, we consider both ViT-Base and ViT-Large variants in order to study the consistency of the proposed textual supervision strategies across different backbone scales. For RemoteCLIP, we further evaluate the ResNet-50 variant, thereby extending the comparison to a convolutional visual encoder. Throughout all experiments, both the visual and text encoders are kept frozen, such that performance differences can be attributed solely to the proposed description generation, aggregation, and calibration strategies rather than to any task-specific adaptation of the backbone models.

\paragraph{Evaluated LLMs} For LLM-based prompt generation, we employ a diverse set of open-source large language models (LLMs) with different architectural characteristics, parameter scales, and levels of domain specialization. In particular, we evaluate \llama~\cite{llama3_1} with 8B parameters, \deepseek~\cite{bi2024deepseek}, \qwen~\cite{qwen2_5}, and the domain-specific \rsllava~\cite{bazi2024rs}, each in their 7B configuration, together with \mixtral~\cite{mixtral}, which adopts an 8$\times$7B mixture-of-experts design. Each LLM is used within the same meta-prompting pipeline to generate class-specific textual descriptions for every target dataset and for all classes in its label space. More specifically, for each dataset-specific query and each class, we generate four independent textual descriptions, and we keep this repetition factor fixed across all evaluated LLMs and datasets. To ensure a controlled comparison, we use a standard temperature of 0.7 and a maximum generation length of 70 tokens for all models, except for \rsllava, where a higher temperature of 1.05 is used to encourage greater descriptive diversity. This fixed generation setup allows us to study how architectural and domain-specific differences among LLMs influence the resulting text descriptions and their effectiveness in downstream zero-shot classification and retrieval.
\section{Quantitative Results}
\label{sec:quant_results}

\subsection{Classification}
\label{sub:classficiation}
\begin{table*}[t]
\centering
\caption{Zero-shot classification accuracy (\%) on all evaluated remote sensing datasets under the Domain-Specific Template (\dst) baseline. Results are reported across the full set of vision-language models and backbone architectures. Best results per dataset are in \textbf{bold}, second best are \underline{underlined}.}
\label{tab:cls_dst}
\resizebox{\textwidth}{!}{
\begin{tabular}{lcccccccccccccc}
\toprule
\SC{Model} & \SC{Arch}
& \rotatebox{90}{\eurosat}
& \rotatebox{90}{\aid}
& \rotatebox{90}{\mlrsnet}
& \rotatebox{90}{\optimal}
& \rotatebox{90}{\patternnet}
& \rotatebox{90}{\resisc}
& \rotatebox{90}{\rsc}
& \rotatebox{90}{\rsicblow}
& \rotatebox{90}{\rsicbhigh}
& \rotatebox{90}{\whurs}
& \rotatebox{90}{\siriwhu}
& \rotatebox{90}{\rsscn}
& \rotatebox{90}{\SC{Avg}} \\
\midrule
\multirow{2}{*}{CLIP} & ViT-B/16 & 51.4 & 67.3 & 57.2 & 72.2 & 64.5 & 64.4 & 62.4 & 30.9 & 43.0 & 85.5 & 55.3 & 65.4 & 59.9 \\
& ViT-L/14 & 63.3 & 69.4 & 65.6 & 80.7 & 75.9 & 71.6 & 66.2 & 40.2 & 48.9 & 85.5 & 65.2 & 64.2 & 66.4 \\
\midrule
\multirow{2}{*}{MetaCLIP-400M} & ViT-B/16 & 53.4 & 71.8 & 59.8 & 77.7 & 69.1 & 70.0 & 65.3 & 33.6 & 39.4 & 90.0 & 60.5 & 67.1 & 63.1 \\
& ViT-L/14 & 50.1 & 73.9 & 62.7 & 80.9 & 74.4 & 71.9 & 64.7 & 35.5 & 44.8 & 89.2 & 66.5 & 69.4 & 65.3 \\
\midrule
\multirow{2}{*}{MetaCLIP-2.5B} & ViT-B/16 & 49.1 & 72.5 & 64.4 & 76.0 & 71.9 & 70.5 & 65.0 & 33.2 & 41.1 & 90.7 & \underline{66.8} & 68.6 & 64.1 \\
& ViT-L/14 & 60.8 & 78.9 & \textbf{70.7} & 83.2 & \textbf{82.8} & \underline{75.0} & 63.1 & 39.5 & 44.0 & 90.0 & 65.8 & 68.9 & 68.6 \\
\midrule
\multirow{2}{*}{SigLIP} & ViT-B/16 & 43.3 & 71.0 & 56.7 & 75.9 & 64.2 & 65.6 & 62.1 & 26.2 & 36.8 & 85.0 & 62.5 & 65.0 & 59.5 \\
& ViT-L/16 & 52.8 & 72.2 & 60.8 & 83.1 & 65.5 & 69.3 & 63.4 & 30.3 & 41.4 & 90.7 & 63.1 & 64.5 & 63.1 \\
\midrule
\multirow{2}{*}{SigLIP2} & ViT-B/16 & 49.2 & 72.7 & 62.5 & 83.1 & 71.0 & 71.2 & 61.1 & 34.5 & 42.7 & 90.8 & 60.5 & 68.1 & 64.0 \\
& ViT-L/16 & 54.5 & 74.7 & \underline{70.2} & \underline{85.3} & 78.9 & 75.0 & 70.5 & 39.8 & 44.4 & 93.0 & 62.9 & 70.3 & 68.3 \\
\midrule
\multirow{3}{*}{RemoteCLIP} & RN-50 & 23.2 & \underline{89.4} & 42.3 & 63.8 & 45.9 & 51.4 & 67.1 & 16.3 & 35.0 & 92.8 & 54.0 & 49.5 & 52.5 \\
& ViT-B/32 & 34.5 & \textbf{91.5} & 56.9 & 77.5 & 59.3 & 67.9 & 62.6 & 26.2 & 39.7 & \textbf{95.1} & 63.9 & 59.8 & 61.2 \\
& ViT-L/14 & 42.9 & 84.2 & 61.6 & 82.5 & 64.1 & \textbf{75.4} & 62.3 & 33.9 & 49.3 & 90.2 & \textbf{67.0} & 64.9 & 64.9 \\
\midrule
\multirow{2}{*}{GeoRSCLIP} & ViT-B/32 & 47.6 & 72.1 & 64.1 & 79.8 & 76.6 & 69.2 & 66.4 & 29.9 & 46.4 & 91.3 & 60.3 & \textbf{77.0} & 65.1 \\
& ViT-L/14 & 64.3 & 76.9 & 67.8 & \textbf{86.5} & \underline{80.9} & 73.9 & \textbf{78.1} & 41.4 & \textbf{55.7} & \underline{94.1} & 63.7 & \underline{72.4} & \textbf{71.3} \\
\midrule
CLIP LAION-RS & ViT-L/14 & \underline{68.8} & 72.5 & 66.1 & 82.4 & 77.8 & 74.1 & 68.4 & \underline{42.3} & 50.4 & 89.0 & 66.6 & 68.7 & 68.9 \\
\midrule
SkyCLIP & ViT-L/14 & \textbf{69.7} & 72.2 & 68.3 & 83.5 & 80.6 & 73.2 & \underline{74.9} & \textbf{48.5} & \underline{53.1} & 89.2 & 66.5 & 69.1 & \underline{70.7} \\
\bottomrule
\end{tabular}
}
\end{table*}
\begin{table*}[ht!]
\centering
\caption{Zero-shot classification accuracy (\%) on all evaluated remote sensing datasets under the meta-prompting approach with \llama. Results are reported across the full set of vision-language models and backbone architectures. Best results per dataset are in \textbf{bold}, second best are \underline{underlined}.}
\label{tab:cls_llama}
\resizebox{\textwidth}{!}{
\begin{tabular}{lcccccccccccccc}
\toprule
\SC{Model} & \SC{Arch}
& \rotatebox{90}{\eurosat}
& \rotatebox{90}{\aid}
& \rotatebox{90}{\mlrsnet}
& \rotatebox{90}{\optimal}
& \rotatebox{90}{\patternnet}
& \rotatebox{90}{\resisc}
& \rotatebox{90}{\rsc}
& \rotatebox{90}{\rsicblow}
& \rotatebox{90}{\rsicbhigh}
& \rotatebox{90}{\whurs}
& \rotatebox{90}{\siriwhu}
& \rotatebox{90}{\rsscn}
& \rotatebox{90}{\SC{Avg}} \\
\midrule
\multirow{2}{*}{CLIP} & ViT-B/16 & 52.5 & 69.3 & 56.5 & 74.5 & 66.6 & 65.8 & 63.6 & 28.6 & 37.2 & 86.3 & 52.9 & 70.1 & 60.3 \\
& ViT-L/14 & 59.2 & 70.3 & 62.7 & 78.5 & 74.6 & 69.5 & 61.8 & 32.7 & 41.2 & 88.2 & 57.8 & 67.8 & 63.7 \\
\midrule
\multirow{2}{*}{MetaCLIP-400M} & ViT-B/16 & 54.0 & 67.8 & 59.3 & 79.3 & 70.7 & 68.6 & 70.1 & 34.8 & 41.7 & 87.2 & 57.9 & 73.5 & 63.7 \\
& ViT-L/14 & 55.4 & 71.4 & 62.6 & 79.7 & 77.1 & 71.5 & 68.9 & 30.8 & 41.0 & 89.5 & 62.6 & 74.2 & 65.4 \\
\midrule
\multirow{2}{*}{MetaCLIP-2.5B} & ViT-B/16 & 44.9 & 69.2 & 62.3 & 78.8 & 73.4 & 70.5 & 66.7 & 35.4 & 40.0 & 89.4 & 57.3 & 74.2 & 63.5 \\
& ViT-L/14 & 59.4 & 74.6 & 66.7 & 83.8 & \underline{80.6} & 73.2 & 66.4 & 40.7 & 39.0 & \underline{90.9} & 62.3 & 72.6 & 67.5 \\
\midrule
\multirow{2}{*}{SigLIP} & ViT-B/16 & 38.5 & 69.9 & 58.3 & 76.5 & 66.2 & 65.8 & 62.9 & 30.3 & 41.5 & 83.3 & 55.6 & 67.1 & 59.7 \\
& ViT-L/16 & 49.3 & 76.1 & 64.7 & 81.0 & 71.3 & 72.0 & \underline{72.9} & 33.2 & 48.7 & 89.8 & 58.3 & 71.7 & 65.7 \\
\midrule
\multirow{2}{*}{SigLIP2} & ViT-B/16 & 41.7 & 71.8 & 62.5 & 80.9 & 72.1 & 70.3 & 64.2 & 33.2 & 44.8 & 87.4 & 57.2 & 69.1 & 62.9 \\
& ViT-L/16 & 55.0 & 76.2 & 68.1 & 83.1 & 76.8 & \underline{74.4} & 72.8 & 37.8 & 47.9 & 91.7 & 60.4 & 71.1 & 67.9 \\
\midrule
\multirow{3}{*}{RemoteCLIP} & RN-50 & 26.1 & \underline{81.9} & 45.2 & 69.5 & 48.8 & 56.5 & 62.3 & 22.0 & 31.8 & \underline{92.9} & 54.7 & 65.7 & 54.8 \\
& ViT-B/32 & 38.0 & \textbf{87.3} & 58.2 & 79.9 & 59.5 & 72.5 & 65.5 & 27.7 & 39.0 & \textbf{95.3} & \underline{68.3} & 71.6 & 63.6 \\
& ViT-L/14 & 45.8 & 80.9 & 60.7 & 81.2 & 63.6 & 74.1 & 67.9 & 31.4 & \underline{51.0} & 91.2 & \textbf{72.6} & 68.0 & 65.7 \\
\midrule
\multirow{2}{*}{GeoRSCLIP} & ViT-B/32 & 52.8 & 72.0 & 65.7 & 80.6 & 78.0 & 71.0 & 72.7 & 34.1 & 48.6 & 89.9 & 60.6 & \textbf{79.7} & 67.1 \\
& ViT-L/14 & \underline{69.2} & 74.5 & \textbf{70.2} & \textbf{84.9} & 80.0 & \textbf{74.6} & \textbf{74.4} & \underline{41.3} & \textbf{53.5} & 90.6 & 61.1 & \underline{77.5} & \textbf{71.0} \\
\midrule
CLIP LAION-RS & ViT-L/14 & \textbf{69.9} & 70.7 & 63.5 & 79.0 & 74.6 & 71.3 & 65.1 & 39.8 & 48.9 & 89.5 & 60.8 & 72.8 & 67.2 \\
\midrule
SkyCLIP & ViT-L/14 & 68.5 & 72.5 & \underline{68.9} & \underline{84.1} & \textbf{82.1} & 74.0 & 72.5 & \textbf{43.9} & 50.8 & 90.9 & 61.1 & 72.2 & \underline{70.1} \\
\bottomrule
\end{tabular}
}
\end{table*}
\paragraph{Vanilla Embeddings} We report results on the zero-shot image classification task, using top-1 accuracy as the evaluation metric. We focus here on the Domain-Specific templates (\dst) and on LLaMA-generated descriptions (\llama) approaches in Table~\ref{tab:cls_dst} and Table~\ref{tab:cls_llama} respectively. \llama is selected as the representative LLM-based configuration because it proved to be the most consistent choice among the evaluated open-source LLMs throughout our experiments. Tables~\ref{tab:cls_st}--\ref{tab:cls_rsllava} in the Appendix, present the corresponding results for the Simple Template (\st) and the remaining LLMs (\deepseek, \qwen, \mixtral, \rsllava) across all models and datasets.

From the \textit{model perspective}, remote sensing-adapted VLMs consistently achieve stronger performance than general-purpose models, regardless of the description generation strategy. This trend is especially pronounced for recent domain-specialized models such as GeoRSCLIP and SkyCLIP, which deliver the most consistent results. Among the general-purpose VLMs, SigLIP2 and MetaCLIP stand out as the most competitive alternatives.

From the \textit{description-generation perspective}, the template-based baselines provide strong and reliable reference points. In particular, \dst consistently outperforms \st, highlighting the importance of using diverse remote-sensing-aware templates rather than a single generic formulation. On the contrary, the LLM-based strategy does not yield a consistent improvement over template-based descriptions. In fact, it leads to performance degradation for the majority of dataset-model combinations. This behavior can be attributed to the fact that, although LLM-generated descriptions introduce semantically richer and more diverse information, they also introduce variability that is not always aligned with the discriminative structure of the downstream visual categories. In practice, these descriptions may contain overly generic attributes, stylistic variation, or semantically adjacent but discriminatively irrelevant details, which can blur the resulting class prototypes in the shared embedding space. As a result, the additional semantic richness of LLM-generated text does not necessarily translate into better alignment with visual representations, and often acts as a source of noise for zero-shot classification.

\input{plots/bars}
%\vspace{-25pt}
Figure~\ref{fig:bars} provides a complementary view to the aforementioned classification results by reporting, for each dataset, the performance of GeoRSCLIP ViT-L/14 model, across all evaluated description-generation strategies. For the LLM-based approaches, each score corresponds to the average classification accuracy obtained from three independently generated description sets per LLM, while the error bars denote the associated standard deviation. This protocol is adopted to account for the intrinsic stochasticity of LLM-based generation and to provide a more reliable estimate of the performance variability induced by different sampled description sets. 

Although the aggregated results indicate that template-based strategies generally outperform LLM-based ones on average across datasets, Figure~\ref{fig:bars} shows that this behavior is dataset-dependent. In particular, the number of cases in which \dst clearly surpasses all LLM-based approaches, is relatively limited (e.g., \rsc). For the majority of datasets, the best-performing description source varies across LLMs, indicating that no single LLM consistently dominates across benchmarks. This dataset-specific behavior is illustrated by several contrasting examples. \qwen for instance, achieves the highest performance on \rsicblow and \rsscn, while yielding the weakest results on \patternnet, \whurs, and \siriwhu. This instability is further reflected in the relatively large standard deviations observed for several LLMs, most notably \rsllava, suggesting that the effectiveness of meta-prompting is highly contingent on the dataset and prompt formulation, in contrast to the more stable behavior exhibited by template-based formulations.

%\vspace{-25pt}

\begin{table*}[t!]
\centering
\caption{Effect of image-side embedding centering on average zero-shot classification accuracy (\%) across all evaluated datasets and description-generation strategies. Reported accuracies correspond to the centered setting, while $\Delta$ denotes the relative performance change with respect to the vanilla, non-centered, image embeddings. Positive values indicate gains and negative values indicate degradation.}
\label{tab:cls_centering}
\resizebox{\textwidth}{!}{
\begin{tabular}{lccc|cc|cc|cc|cc|cc|cc}
\toprule
\multirow{2}{*}{\textbf{\SC{Model}}} & \multirow{2}{*}{\textbf{\SC{Arch}}}
& \multicolumn{2}{c}{\textbf{\st}}
& \multicolumn{2}{c}{\textbf{\dst}}
& \multicolumn{2}{c}{\textbf{\llama}}
& \multicolumn{2}{c}{\textbf{\deepseek}}
& \multicolumn{2}{c}{\textbf{\qwen}}
& \multicolumn{2}{c}{\textbf{\mixtral}}
& \multicolumn{2}{c}{\textbf{\rsllava}} \\
\cmidrule(lr){3-16}
& & centered & $\Delta$
& centered & $\Delta$
& centered & $\Delta$
& centered & $\Delta$
& centered & $\Delta$
& centered & $\Delta$
& centered & $\Delta$ \\
\midrule

\multirow{2}{*}{CLIP} & ViT-B/16 & 62.1 &\textcolor{ForestGreen}{+4.8} & 62.5 &\textcolor{ForestGreen}{+2.5} & 63.7 &\textcolor{ForestGreen}{+3.4} & 63.9 &\textcolor{ForestGreen}{+4.3} & 63.9 &\textcolor{ForestGreen}{+4.6} & 62.3 &\textcolor{ForestGreen}{+3.0} & 62.4 &\textcolor{ForestGreen}{+3.9} \\
& ViT-L/14 & 66.4 &\textcolor{ForestGreen}{+2.3} & 66.7 &\textcolor{ForestGreen}{+0.3} & 68.5 &\textcolor{ForestGreen}{+4.8} & 68.2 &\textcolor{ForestGreen}{+3.9} & 69.1 &\textcolor{ForestGreen}{+4.6} & 67.7 &\textcolor{ForestGreen}{+2.3} & 68.1 &\textcolor{ForestGreen}{+4.0} \\
\midrule

\multirow{2}{*}{MetaCLIP-400M} & ViT-B/16 & 64.5 &\textcolor{ForestGreen}{+2.2} & 65.7 &\textcolor{ForestGreen}{+2.5} & 66.3 &\textcolor{ForestGreen}{+2.6} & 65.8 &\textcolor{ForestGreen}{+3.2} & 66.5 &\textcolor{ForestGreen}{+4.1} & 64.2 &\textcolor{ForestGreen}{+3.2} & 64.3 &\textcolor{ForestGreen}{+2.7} \\
& ViT-L/14 & 65.6 &\textcolor{ForestGreen}{+2.3} & 67.4 &\textcolor{ForestGreen}{+2.1} & 68.3 &\textcolor{ForestGreen}{+2.9} & 66.9 &\textcolor{ForestGreen}{+3.1} & 68.2 &\textcolor{ForestGreen}{+3.6} & 66.2 &\textcolor{ForestGreen}{+1.8} & 66.2 &\textcolor{ForestGreen}{+2.8} \\
\midrule

\multirow{2}{*}{MetaCLIP-2.5B} & ViT-B/16 & 65.7 &\textcolor{ForestGreen}{+3.9} & 65.6 &\textcolor{ForestGreen}{+1.5} & 66.5 &\textcolor{ForestGreen}{+3.0} & 65.9 &\textcolor{ForestGreen}{+4.2} & 66.5 &\textcolor{ForestGreen}{+3.7} & 64.9 &\textcolor{ForestGreen}{+3.7} & 64.8 &\textcolor{ForestGreen}{+2.0} \\
& ViT-L/14 & 71.2 &\textcolor{ForestGreen}{+3.5} & 71.1 &\textcolor{ForestGreen}{+2.6} & 71.3 &\textcolor{ForestGreen}{+3.8} & 70.5 &\textcolor{ForestGreen}{+3.9} & 71.4 &\textcolor{ForestGreen}{+5.0} & 69.1 &\textcolor{ForestGreen}{+3.6} & 69.9 &\textcolor{ForestGreen}{+3.6} \\
\midrule

\multirow{2}{*}{SigLIP} & ViT-B/16 & 58.9 & \textcolor{red}{-0.6} & 59.1 & \textcolor{red}{-0.4} & 61.8 &\textcolor{ForestGreen}{+2.2} & 61.3 &\textcolor{ForestGreen}{+2.5} & 61.4 &\textcolor{ForestGreen}{+3.2} & 60.0 &\textcolor{ForestGreen}{+2.7} & 60.9 &\textcolor{ForestGreen}{+1.7} \\
& ViT-L/16 & 65.3 &\textcolor{ForestGreen}{+2.0} & 65.0 &\textcolor{ForestGreen}{+1.9} & 67.9 &\textcolor{ForestGreen}{+2.2} & 67.6 &\textcolor{ForestGreen}{+3.1} & 67.4 &\textcolor{ForestGreen}{+2.4} & 66.1 &\textcolor{ForestGreen}{+3.8} & 67.1 &\textcolor{ForestGreen}{+2.5} \\
\midrule
\multirow{2}{*}{SigLIP2} & ViT-B/16 & 65.4 &\textcolor{ForestGreen}{+1.9} & 65.4 &\textcolor{ForestGreen}{+1.4} & 65.3 &\textcolor{ForestGreen}{+2.4} & 65.1 &\textcolor{ForestGreen}{+2.9} & 65.1 &\textcolor{ForestGreen}{+2.6} & 63.7 &\textcolor{ForestGreen}{+3.1} & 64.0 &\textcolor{ForestGreen}{+1.2} \\
& ViT-L/16 & 70.5 &\textcolor{ForestGreen}{+0.7} & 70.0 &\textcolor{ForestGreen}{+1.7} & 71.2 &\textcolor{ForestGreen}{+3.3} & 70.2 &\textcolor{ForestGreen}{+3.5} & 70.9 &\textcolor{ForestGreen}{+4.0} & 69.0 &\textcolor{ForestGreen}{+2.8} & 69.4 &\textcolor{ForestGreen}{+1.9} \\
\midrule

\multirow{3}{*}{RemoteCLIP} & RN-50 & 53.3 &\textcolor{ForestGreen}{+1.9} & 55.5 &\textcolor{ForestGreen}{+3.0} & 57.1 &\textcolor{ForestGreen}{+2.3} & 57.2 &\textcolor{ForestGreen}{+1.0} & 56.1 &\textcolor{ForestGreen}{+1.6} & 57.0 &\textcolor{ForestGreen}{+1.0} & 55.6 &\textcolor{ForestGreen}{+2.0} \\
& ViT-B/32 & 61.8 &\textcolor{ForestGreen}{+1.2} & 63.1 &\textcolor{ForestGreen}{+1.9} & 64.8 &\textcolor{ForestGreen}{+1.3} & 64.9 &\textcolor{ForestGreen}{+1.1} & 64.0 &\textcolor{ForestGreen}{+1.4} & 64.7 & \textcolor{red}{-0.1} & 64.5 &\textcolor{ForestGreen}{+0.8} \\
& ViT-L/14 & 66.3 &\textcolor{ForestGreen}{+1.9} & 67.2 &\textcolor{ForestGreen}{+2.3} & 68.8 &\textcolor{ForestGreen}{+3.1} & 68.3 &\textcolor{ForestGreen}{+3.7} & 68.7 &\textcolor{ForestGreen}{+3.0} & 67.4 &\textcolor{ForestGreen}{+2.4} & 67.6 &\textcolor{ForestGreen}{+3.4} \\
\midrule

\multirow{2}{*}{GeoRSCLIP} & ViT-B/32 & 66.5 &\textcolor{ForestGreen}{+2.2} & 66.3 &\textcolor{ForestGreen}{+1.3} & 69.0 &\textcolor{ForestGreen}{+1.9} & 67.9 &\textcolor{ForestGreen}{+1.1} & 68.8 &\textcolor{ForestGreen}{+2.9} & 66.3 & \textcolor{red}{-0.1} & 66.8 &\textcolor{ForestGreen}{+0.5} \\
& ViT-L/14 & 71.7 &\textcolor{ForestGreen}{+0.5} & 71.3 &\textcolor{ForestGreen}{+0.0} & 74.1 &\textcolor{ForestGreen}{+3.1} & 73.7 &\textcolor{ForestGreen}{+2.8} & 74.0 &\textcolor{ForestGreen}{+3.7} & 72.1 &\textcolor{ForestGreen}{+0.5} & 73.5 &\textcolor{ForestGreen}{+2.0} \\
\midrule
CLIP LAION-RS & ViT-L/14 & 68.6 &\textcolor{ForestGreen}{+0.1} & 68.7 & \textcolor{red}{-0.3} & 70.2 &\textcolor{ForestGreen}{+3.0} & 69.4 &\textcolor{ForestGreen}{+2.4} & 70.7 &\textcolor{ForestGreen}{+3.0} & 68.9 &\textcolor{ForestGreen}{+1.9} & 69.7 &\textcolor{ForestGreen}{+2.6} \\
\midrule
SkyCLIP & ViT-L/14 & 69.4 & \textcolor{red}{-1.2} & 69.5 & \textcolor{red}{-1.2} & 71.0 &\textcolor{ForestGreen}{+0.9} & 70.5 & \textcolor{red}{-0.1} & 71.5 &\textcolor{ForestGreen}{+0.7} & 70.2 & \textcolor{red}{-0.6} & 71.1 &\textcolor{ForestGreen}{+0.4} \\

\bottomrule
\end{tabular}
}
\vspace{-15pt}
\end{table*}
\paragraph{Embedding Calibration}
Table~\ref{tab:cls_centering} reports the effect of image-side embedding centering in the zero-shot classification task, averaged across all evaluated datasets. For each vision-language model and description-generation strategy, we report both the classification accuracy obtained after centering and the corresponding performance change ($\mathbf{\Delta}$) relative to the vanilla setting. 
% Since classification is performed by matching each image embedding against the set of class prototypes, the query modality in this case is the image. Accordingly, calibration is applied by subtracting the mean image embedding prior to similarity computation.

The results indicate that image centering constitutes a simple yet highly effective calibration step (as shown in Figure~\ref{fig:placeholder}). Performance gains are widespread, reaching up to $\mathbf{+4.8\%}$. At the same time, performance degradation is relatively rare and is mostly confined to template-based descriptions. For example, in the case of SkyCLIP, both \st and \dst decrease by approximately $\mathbf{1.2\%}$, while \deepseek and \mixtral exhibit smaller drops of about $\mathbf{0.1\%}$ and $\mathbf{0.6\%}$, respectively. In contrast, \llama, \qwen, and \rsllava all benefit from centering. This pattern indicates that image centering improves alignment in a highly consistent manner, with only a few isolated exceptions.

An important observation is that embedding calibration changes the relative ranking between template-based and LLM-based descriptions. Feature centering acts as a significant enabler for high-capacity models like \llama and \qwen (and \deepseek to a lesser degree), making the LLM-generated descriptions consistently surpass \st and \dst. This suggests that a considerable part of the underperformance of meta-prompting in the vanilla setting depends on how their richer and more variable semantics interact with the uncalibrated embedding space. Centering, on the other hand, effectively improves the alignment between image embeddings and the more diverse textual prototypes produced by LLMs. We note that \mixtral and \rsllava lag behind other LLMs even post-calibration and frequently trail the simple \st baseline (e.g., under SigLIP2 ViT-L/16), suggesting an inherent stylistic or structural incompatibility that geometric centering alone cannot resolve.

Reading the table column-wise shifts the focus to architectural capacity and highlights the definitive advantage of remote-sensing-adapted VLMs over general-purpose alternatives. Within any given prompting strategy or calibration state, domain-specific architectures like RemoteCLIP, GeoRSCLIP, and SkyCLIP consistently outperform standard baselines like CLIP or MetaCLIP. This margin is most pronounced for larger baselines. For instance, under the calibrated \qwen text generation, GeoRSCLIP ViT-L/14 achieves $74.0\%$, a $4.9\%$ improvement over the general-purpose CLIP ViT-L/14 counterpart ($69.1\%$). Overall, the results suggest that pre-training on domain-specific imagery establishes a fundamentally more resilient embedding geometry for downstream zero-shot tasks.

\subsection{Text-to-Image Retrieval}
In this section, we evaluate text-to-image retrieval task, using the mean Average Precision (mAP) metric. Table~\ref{tab:ret_centering} summarizes the average retrieval performance across the evaluated datasets, for every vision-language model and description-generation strategy. We report results both before and after text-side embedding centering, together with the corresponding performance differences ($\mathbf{\Delta}$). In this task, each aggregated class prototype is used as a text query to rank the image gallery. Therefore, the query modality corresponds to the text side.
\begin{table*}[ht!]
\centering
\caption{Effect of text-side embedding centering on average text-to-image retrieval performance across all evaluated datasets. We report the mAP metric before (\textit{mAP}) and after (\textit{mAP$_c$}) calibration for each description-generation strategy. $\Delta$ denotes the relative performance change with respect to the vanilla, non-centered, text embeddings.}
\label{tab:ret_centering}
\resizebox{\textwidth}{!}{
\begin{tabular}{lcccc|ccc|ccc|ccc|ccc|ccc|ccc}
\toprule
\multirow{2}{*}{\textbf{\SC{Model}}} &
\multirow{2}{*}{\textbf{\SC{Arch}}}
& \multicolumn{3}{c}{\textbf{\st}}
& \multicolumn{3}{c}{\textbf{\dst}}
& \multicolumn{3}{c}{\textbf{\llama}}
& \multicolumn{3}{c}{\textbf{\deepseek}}
& \multicolumn{3}{c}{\textbf{\qwen}}
& \multicolumn{3}{c}{\textbf{\mixtral}}
& \multicolumn{3}{c}{\textbf{\rsllava}} \\
\cmidrule(lr){3-23}
& &
mAP & mAP$_c$ & $\Delta$ &
mAP & mAP$_c$  & $\Delta$ &
mAP & mAP$_c$ & $\Delta$ &
mAP & mAP$_c$ & $\Delta$ &
mAP & mAP$_c$ & $\Delta$ &
mAP & mAP$_c$ & $\Delta$ &
mAP & mAP$_c$ & $\Delta$ \\
\midrule

\multirow{2}{*}{CLIP} & ViT-B/16
& 51.1 & 62.5 &\textcolor{ForestGreen}{+11.4}
& 52.9 & 63.5 &\textcolor{ForestGreen}{+10.6}
& 51.8 & 63.9 &\textcolor{ForestGreen}{+12.1}
& 51.0 & 64.6 &\textcolor{ForestGreen}{+13.6}
& 50.9 & 63.7 &\textcolor{ForestGreen}{+12.8} 
& 52.7 & 64.4 &\textcolor{ForestGreen}{+11.7}
& 50.0 & 64.3 &\textcolor{ForestGreen}{+14.3} \\

& ViT-L/14
& 55.4 & 68.0 &\textcolor{ForestGreen}{+12.6}
& 59.2 & 68.4 &\textcolor{ForestGreen}{+9.2}
& 54.0 & 69.0 &\textcolor{ForestGreen}{+15.0}
& 53.5 & 69.5 &\textcolor{ForestGreen}{+16.0}
& 53.6 & 68.8 &\textcolor{ForestGreen}{+15.2}
& 55.4 & 69.4 &\textcolor{ForestGreen}{+14.0}
& 52.7 & 69.7 &\textcolor{ForestGreen}{+17.0} \\
\midrule

\multirow{2}{*}{MetaCLIP-400M} & ViT-B/16
& 55.6 & 64.7 &\textcolor{ForestGreen}{+9.1}
& 59.3 & 65.2 &\textcolor{ForestGreen}{+5.9}
& 55.6 & 65.8 &\textcolor{ForestGreen}{+10.2}
& 56.0 & 66.5 &\textcolor{ForestGreen}{+10.5}
& 54.5 & 65.5 &\textcolor{ForestGreen}{+11.0}
& 58.6 & 65.5 &\textcolor{ForestGreen}{+6.9}
& 55.0 & 66.2 &\textcolor{ForestGreen}{+11.2} \\

& ViT-L/14
& 56.9 & 65.9 &\textcolor{ForestGreen}{+9.0}
& 58.5 & 66.9 &\textcolor{ForestGreen}{+8.4}
& 55.5 & 68.0 &\textcolor{ForestGreen}{+12.5}
& 55.4 & 68.3 &\textcolor{ForestGreen}{+12.9}
& 54.6 & 67.8 &\textcolor{ForestGreen}{+13.2}
& 57.0 & 67.6 &\textcolor{ForestGreen}{+10.6}
& 53.9 & 68.0 &\textcolor{ForestGreen}{+14.1} \\
\midrule

\multirow{2}{*}{MetaCLIP-2.5B} & ViT-B/16
& 57.6 & 66.4 &\textcolor{ForestGreen}{+8.8}
& 61.5 & 66.5 &\textcolor{ForestGreen}{+5.0}
& 57.3 & 67.0 &\textcolor{ForestGreen}{+9.7}
& 57.3 & 67.6 &\textcolor{ForestGreen}{+10.3}
& 55.8 & 66.5 &\textcolor{ForestGreen}{+10.7}
& 60.2 & 67.2 &\textcolor{ForestGreen}{+7.0}
& 56.7 & 67.4 &\textcolor{ForestGreen}{+10.7} \\

& ViT-L/14
& 62.2 & 70.8 &\textcolor{ForestGreen}{+8.6}
& 63.5 & 71.0 &\textcolor{ForestGreen}{+7.5}
& 59.7 & 71.2 &\textcolor{ForestGreen}{+11.5}
& 59.6 & 71.9 &\textcolor{ForestGreen}{+12.3}
& 58.7 & 70.6 &\textcolor{ForestGreen}{+11.9}
& 62.5 & 71.0 &\textcolor{ForestGreen}{+8.5}
& 58.8 & 71.2 &\textcolor{ForestGreen}{+12.4} \\
\midrule

\multirow{2}{*}{SigLIP} & ViT-B/16
& 49.6 & 61.0 & \textcolor{ForestGreen}{+11.4}
& 54.9 & 61.2 & \textcolor{ForestGreen}{+6.3}
& 50.8 & 61.6 &\textcolor{ForestGreen}{+10.8}
& 50.8 & 62.1 &\textcolor{ForestGreen}{+11.3}
& 49.8 & 61.6 &\textcolor{ForestGreen}{+11.8}
& 52.9 & 61.5 &\textcolor{ForestGreen}{+8.6}
& 49.5 & 62.1 &\textcolor{ForestGreen}{+12.6} \\

& ViT-L/16
& 54.7 & 65.7 &\textcolor{ForestGreen}{+11.0}
& 58.6 & 65.9 &\textcolor{ForestGreen}{+7.3}
& 56.0 & 67.5 &\textcolor{ForestGreen}{+11.5}
& 55.8 & 67.9 &\textcolor{ForestGreen}{+12.1}
& 55.1 & 67.3 &\textcolor{ForestGreen}{+12.2}
& 57.2 & 67.0 &\textcolor{ForestGreen}{+9.8}
& 54.9 & 62.6 &\textcolor{ForestGreen}{+7.7} \\
\midrule

\multirow{2}{*}{SigLIP2} & ViT-B/16
& 56.9 & 65.8 &\textcolor{ForestGreen}{+8.9}
& 61.8 & 66.5 &\textcolor{ForestGreen}{+4.7}
& 54.9 & 65.7 &\textcolor{ForestGreen}{+10.8}
& 55.3 & 66.4 &\textcolor{ForestGreen}{+11.1}
& 53.3 & 65.8 &\textcolor{ForestGreen}{+12.5}
& 59.1 & 65.8 &\textcolor{ForestGreen}{+6.7}
& 55.4 & 66.4 &\textcolor{ForestGreen}{+11.0} \\

& ViT-L/16
& 61.1 & 70.5 &\textcolor{ForestGreen}{+9.4}
& 64.1 & 70.4 &\textcolor{ForestGreen}{+6.3}
& 57.8 & 70.9 &\textcolor{ForestGreen}{+13.1}
& 57.5 & 70.8 &\textcolor{ForestGreen}{+13.3}
& 56.5 & 70.7 &\textcolor{ForestGreen}{+14.2}
& 60.2 & 70.0 &\textcolor{ForestGreen}{+9.8}
& 56.8 & 70.9 &\textcolor{ForestGreen}{+14.1} \\
\midrule
 
\multirow{3}{*}{RemoteCLIP} & RN-50
& 43.6 & 56.7 & \textcolor{ForestGreen}{+13.1}
& 52.5 & 58.8 & \textcolor{ForestGreen}{+6.3}
& 41.5 & 57.8 &\textcolor{ForestGreen}{+16.3}
& 41.8 & 60.0 &\textcolor{ForestGreen}{+18.2}
& 38.5 & 57.7 &\textcolor{ForestGreen}{+19.2}
& 45.0 & 60.1 &\textcolor{ForestGreen}{+15.1}
& 42.1 & 58.7 &\textcolor{ForestGreen}{+16.6} \\

& ViT-B/32
& 52.7 & 65.1 &\textcolor{ForestGreen}{+12.4}
& 54.3 & 66.1 &\textcolor{ForestGreen}{+11.8}
& 49.2 & 66.8 &\textcolor{ForestGreen}{+17.6}
& 49.4 & 67.4 &\textcolor{ForestGreen}{+18.0}
& 46.6 & 66.8 &\textcolor{ForestGreen}{+20.2}
& 51.5 & 67.2 &\textcolor{ForestGreen}{+15.7}
& 49.4 & 67.3 &\textcolor{ForestGreen}{+17.9} \\

& ViT-L/14
& 54.3 & 70.0 &\textcolor{ForestGreen}{+15.7}
& 54.5 & 70.3 &\textcolor{ForestGreen}{+15.8}
& 50.9 & 69.8 &\textcolor{ForestGreen}{+18.9}
& 50.9 & 70.5 &\textcolor{ForestGreen}{+19.6}
& 49.9 & 69.6 &\textcolor{ForestGreen}{+19.7}
& 51.8 & 70.3 &\textcolor{ForestGreen}{+18.5}
& 50.5 & 70.2 &\textcolor{ForestGreen}{+19.7} \\
\midrule

\multirow{2}{*}{GeoRSCLIP} & ViT-B/32
& 62.0 & 67.7 &\textcolor{ForestGreen}{+5.7}
& 64.7 & 67.8 &\textcolor{ForestGreen}{+3.1}
& 57.9 & 67.4 &\textcolor{ForestGreen}{+9.5}
& 58.4 & 68.3 &\textcolor{ForestGreen}{+9.9}
& 55.9 & 67.3 &\textcolor{ForestGreen}{+11.4}
& 60.7 & 68.3 &\textcolor{ForestGreen}{+7.6}
& 59.2 & 68.4 &\textcolor{ForestGreen}{+9.2} \\

& ViT-L/14
& \underline{68.1} & \underline{72.4} &\textcolor{ForestGreen}{+4.3}
& \underline{69.8} & \underline{72.6} &\textcolor{ForestGreen}{+2.8}
& \underline{63.3} & \underline{72.2} &\textcolor{ForestGreen}{+8.9}
& \underline{64.2} & \textbf{73.0} &\textcolor{ForestGreen}{+8.8}
& \underline{61.8} & \underline{72.3} &\textcolor{ForestGreen}{+10.5}
& \textbf{67.3} & \underline{73.1} &\textcolor{ForestGreen}{+5.8}
& \underline{65.8} & \textbf{73.7} &\textcolor{ForestGreen}{+7.9} \\
\midrule

CLIP LAION-RS & ViT-B/32
& 65.5 & 71.2 & \textcolor{ForestGreen}{+5.7}
& 67.4 & 71.3 & \textcolor{ForestGreen}{+3.9}
& 62.6 & 70.7 &\textcolor{ForestGreen}{+8.1}
& 62.6 & 71.4 &\textcolor{ForestGreen}{+8.8}
& 61.7 & 70.8 &\textcolor{ForestGreen}{+9.1}
& 65.0 & 71.5 &\textcolor{ForestGreen}{+6.5}
& 63.0 & 71.8 &\textcolor{ForestGreen}{+8.8} \\
\midrule

SkyCLIP & ViT-L/14
& \textbf{68.9} & \textbf{72.7} &\textcolor{ForestGreen}{+3.8}
& \textbf{70.7} & \textbf{72.9} &\textcolor{ForestGreen}{+2.2}
& \textbf{66.5} & \textbf{72.6} &\textcolor{ForestGreen}{+6.1}
& \textbf{66.2} & \underline{73.0} &\textcolor{ForestGreen}{+6.8}
& \textbf{65.9} & \textbf{72.7} &\textcolor{ForestGreen}{+6.8}
& \underline{66.9} & \textbf{73.1} &\textcolor{ForestGreen}{+6.2}
& \textbf{66.4} & \underline{73.4} &\textcolor{ForestGreen}{+7.0} \\

\bottomrule
\end{tabular}
}
\end{table*}

Similarly to the classification task, remote sensing-adapted VLMs remain the most robust performers and consistently outperform general-purpose models across all description-generation strategies. This advantage is even more pronounced in retrieval, where models such as RemoteCLIP, GeoRSCLIP, CLIP-LAION-RS, and SkyCLIP define the strongest performance regime. Among the general-purpose VLMs, the MetaCLIP family emerges as the most competitive alternative. However, a substantial performance gap relative to the domain-adapted models still remains.

Additionally, text-to-image retrieval exhibits substantially larger gains from embedding calibration, with improvements reaching more than $\mathbf{+19\%}$. Crucially, centering does not cause performance degradation for any model or description-generation strategy. This highlights the strong sensitivity of retrieval performance to the geometry of the text embedding space and further confirms the effectiveness and robustness of the proposed calibration step. Both template- and LLM-based descriptions benefit from text-side centering, but the gains are significantly larger for the meta-prompting approaches. This indicates that the richer yet more dispersed semantic structure induced by meta-prompting is especially sensitive to embedding-space biases. By removing dominant shared directions from the text side, centering improves the compatibility between class prototypes and image embeddings, yielding stronger retrieval performance.

\subsection{Experimental Analysis}
\begin{wrapfigure}[20]{r}{0.35\textwidth}
	\vspace{-16pt}
	\centering
	\resizebox{\linewidth}{!}{%
		\begin{tikzpicture}
\begin{axis}[
    width=0.5\textwidth,
    height=0.27\textheight,
    ybar,
    bar width=20pt,
    ymin=-3500,
    ymax=-1500,
    symbolic x coords={DS-Temp,Qwen-2.5,LLaMA-3.1,DeepSeek,RS-LLaVA,Mixtral},
    xtick={DS-Temp,Qwen-2.5,LLaMA-3.1,DeepSeek,RS-LLaVA,Mixtral},
    xticklabels={\dst,\qwen,\llama,\deepseek,\rsllava,\mixtral},
    xticklabel style={font=\scriptsize, rotate=35, anchor=east},
    yticklabel style={font=\scriptsize},
    nodes near coords,
    every node near coord/.append style={
        font=\bfseries\tiny,
        rotate=90,
        anchor=west,
        /pgf/number format/.cd,
        fixed,
        fixed zerofill,
        precision=0,
        /tikz/.cd
    },
    ymajorgrids,
    xmajorgrids=false,
    grid style={gray!20},
    minor y tick num=1,
    yminorgrids=true,
    minor grid style={gray!12},
    enlarge x limits=0.12,
    every axis plot/.append style={draw=none},
]

\addplot[
    mark=none,
    fill=DarkGoldenrod2,
    bar shift=0pt,
    every node near coord/.append style={anchor=east},
] coordinates {(DS-Temp,-1897.21)};

\addplot[
    mark=none,
    fill=DarkOrchid3,
    bar shift=0pt,
    every node near coord/.append style={anchor=east},
] coordinates {(Qwen-2.5,-2480.75)};

\addplot[
    mark=none,
    fill=SteelBlue4,
    bar shift=0pt,
    every node near coord/.append style={anchor=east},
] coordinates {(LLaMA-3.1,-2574.90)};

\addplot[
    mark=none,
    fill=DarkOliveGreen3,
    bar shift=0pt,
    every node near coord/.append style={anchor=east},
] coordinates {(DeepSeek,-2633.28)};

\addplot[
    mark=none,
    fill=IndianRed2,
    bar shift=0pt
] coordinates {(RS-LLaVA,-3073.47)};

\addplot[
    mark=none,
    fill=CadetBlue3,
    bar shift=0pt
] coordinates {(Mixtral,-3302.38)};

\end{axis}
\end{tikzpicture}
	}
	\caption{Average text log-likelihood across all evaluated datasets in the CLIP ViT-L/14 text feature space for the different description-generation strategies. Higher values indicate that the corresponding generated descriptions lie in higher-density regions of the whitened text embedding space.}
	\label{fig:log-like-bars}
\end{wrapfigure}
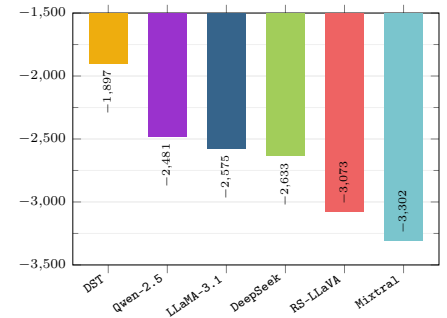
\paragraph{Log-likelihood} To further analyze the geometry of the generated textual descriptions, we compute a text-space log-likelihood score following the formulation of~\cite{betser2026whitened}. Specifically, we evaluate the \dst baseline and all LLM-based meta-prompting strategies by encoding their text descriptions with CLIP ViT-L/14, then applying the same centering and whitening transformation proposed in~\cite{betser2026whitened}.
Let $\mathbf{t} \in \mathbb{R}^D$ denote a text embedding, $\boldsymbol{\mu}$ the reference mean feature vector, and $\mathbf{W}$ the whitening matrix. The whitened text representation is defined as
\begin{equation}
	\tilde{\mathbf{t}} = (\mathbf{t} - \boldsymbol{\mu})\mathbf{W}.
\end{equation}
Assuming a standard Gaussian distribution in the whitened space, the log-likelihood of $\mathbf{t}$ is given by:
\begin{equation}
	\log p(\mathbf{t}) = -\frac{1}{2}\left(D \log(2\pi) + \|\tilde{\mathbf{t}}\|^2\right).
\end{equation}
For the LLM-based approaches, the value shown in Figure~\ref{fig:log-like-bars} corresponds to the average log-likelihood across all evaluated datasets. As shown in Figure~\ref{fig:log-like-bars}, \dst achieves the highest average log-likelihood, whereas all meta-prompting sources produce substantially lower values. This supports the view that, despite their greater semantic richness, LLM-generated descriptions deviate more from the high-density regions of the CLIP text manifold, which is consistent with the noisier and less stable downstream behavior observed in the main evaluation.

% \begin{figure}[t]
	% \centering
	% \input{plots/log_like_bars}
	% \caption{Average text log-likelihood in the text feature space for different prompt sources for CLIP-ViT-L/14. Higher values are better.}
	% \label{fig:log-like-bars}
	% \end{figure}

%\textbf{\begin{figure}[t]
%		\centering
%		\resizebox{0.55\linewidth}{!}{%
%			\input{plots/log_like_bars}
%		}
%		\caption{Average text log-likelihood across all evaluated datasets in the CLIP ViT-L/14 text feature space. Higher values indicate descriptions lying in higher-density regions of the whitened text embedding space.}
%		\label{fig:log-like-bars}
%		\vspace{-8pt}
%	\end{figure}
%}

\begin{figure*}[ht!]
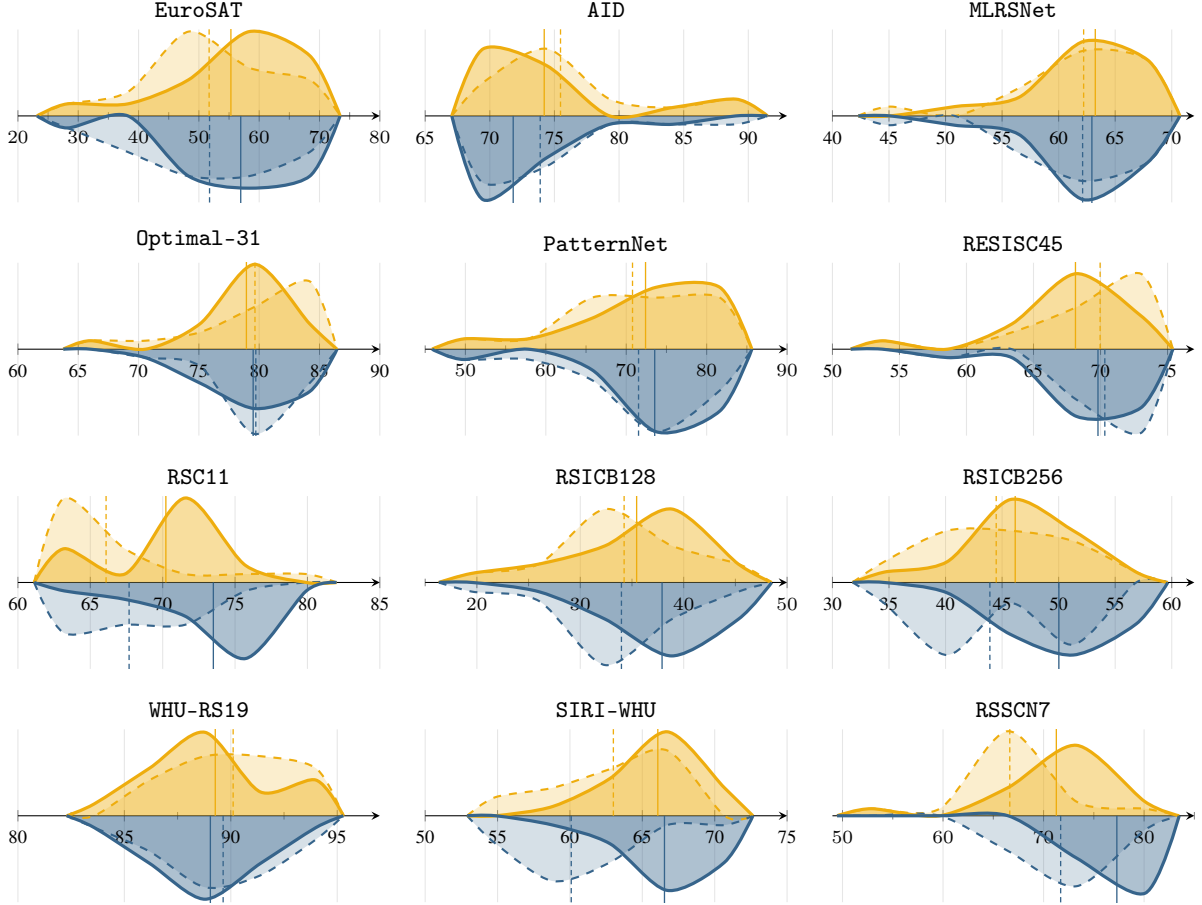

\noindent
\centering
\makebox[\textwidth][c]{%
\begin{tikzpicture}

\begin{groupplot}[
    group style={
        group size=3 by 4,
        horizontal sep=6mm,
        vertical sep=8mm
    },
    width=0.29\textwidth,
    height=0.10\textheight,
    every axis plot/.append style={draw=none},
    scale only axis,
    ymin=-1, ymax=1,
    axis x line=middle,
    axis y line=none,
    ytick=\empty,
    title style={font=\small, yshift=-5pt},
    xmajorgrids,
    % minor x tick num=1,
    % xminorgrids=false,
    xticklabel style={font=\scriptsize},
    grid style={gray!20},
]

\pgfmathsetmacro{\dx}{0}

\nextgroupplot[
    title={\eurosat},
    xmin=20, xmax=80,
    ymin=-0.21, ymax=0.21,
    minor x tick num=1,
    xtick distance=10
]
\input{plots/violin_clas/eurosat.tex}

\nextgroupplot[
    title={\aid},
    xmin=65, xmax=93,
    ymin=-0.62, ymax=0.62,
    xtick={65,70,75,80,85,90},
    xticklabels={65,70,75,80,85,90},
    % xtick distance=10
]
\input{plots/violin_clas/aid.tex}

\nextgroupplot[
    title={\mlrsnet},
    xmin=40, xmax=72,
    ymin=-0.48, ymax=0.48,
    xtick={40,45,50,55,60,65,70},
    xticklabels={40,45,50,55,60,65,70},
    % minor x tick num=1,
    % xtick distance=10
]
\input{plots/violin_clas/mlrsnet.tex}

\nextgroupplot[
    title={\optimal},
    xmin=60, xmax=90,
    ymin=-0.66, ymax=0.66,
    xtick={60,65,70,75,80,85,90},
    xticklabels={60,65,70,75,80,85,90},
    % xtick distance=10
]
\input{plots/violin_clas/optimal31.tex}

\nextgroupplot[
    title={\patternnet},
    xmin=45, xmax=90,
    ymin=-0.31, ymax=0.31,
    minor x tick num=1,
    xtick distance=10
]
\input{plots/violin_clas/patternnet.tex}

\nextgroupplot[
    title={\resisc},
    xmin=50, xmax=77,
    ymin=-0.63, ymax=0.63,
    xtick={50,55,60,65,70,75},
    xticklabels={50,55,60,65,70,75},
]
\input{plots/violin_clas/resisc45.tex}

\nextgroupplot[
    title={\rsc},
    xmin=60, xmax=85,
    ymin=-0.72, ymax=0.72,
    xtick={60,65,70,75,80,85},
    xticklabels={60,65,70,75,80,85},
]
\input{plots/violin_clas/rsc11.tex}

\nextgroupplot[
    title={\rsicblow},
    xmin=15, xmax=50,
    ymin=-0.43, ymax=0.43,
    minor x tick num=1,
    xtick distance=10
]
\input{plots/violin_clas/rsicb128.tex}

\nextgroupplot[
    title={\rsicbhigh},
    xmin=30, xmax=62,
    ymin=-0.44, ymax=0.44,
    % minor x tick num=1,
    % xtick distance=10
    xtick={30,35,40,45,50,55,60},
    xticklabels={30,35,40,45,50,55,60},
]
\input{plots/violin_clas/rsicb256.tex}

\nextgroupplot[
    title={\whurs},
    xmin=80, xmax=97,
    ymin=-0.82, ymax=0.82,
    xtick={80,85,90,95},
    xticklabels={80,85,90,95},
    minor x tick num=1,
]
\input{plots/violin_clas/whurs19.tex}

\nextgroupplot[
    title={\siriwhu},
    xmin=50, xmax=75,
    ymin=-0.69, ymax=0.69,
    xtick={50,55,60,65,70,75},
    xticklabels={50,55,60,65,70,75},
    % xtick distance=10
]
\input{plots/violin_clas/siriwhu.tex}

\nextgroupplot[
    title={\rsscn},
    xmin=49, xmax=85,
    ymin=-0.53, ymax=0.53,
    minor x tick num=1,
    xtick distance=10
]
\input{plots/violin_clas/rsscn7.tex}

\end{groupplot}

\end{tikzpicture}
}
\caption{Dataset-wise fixed-bin histograms of zero-shot classification accuracy, across all evaluated models for \textcolor{\dstempcol}{\dst} and \textcolor{\llamacol}{\llama} generated descriptions. Dashed: results under the non-centered setting, solid: results under the image-side embedding centering.}
\label{fig:violin-cls}
\vspace{-15pt}
\end{figure*}
\begin{figure*}[ht!]
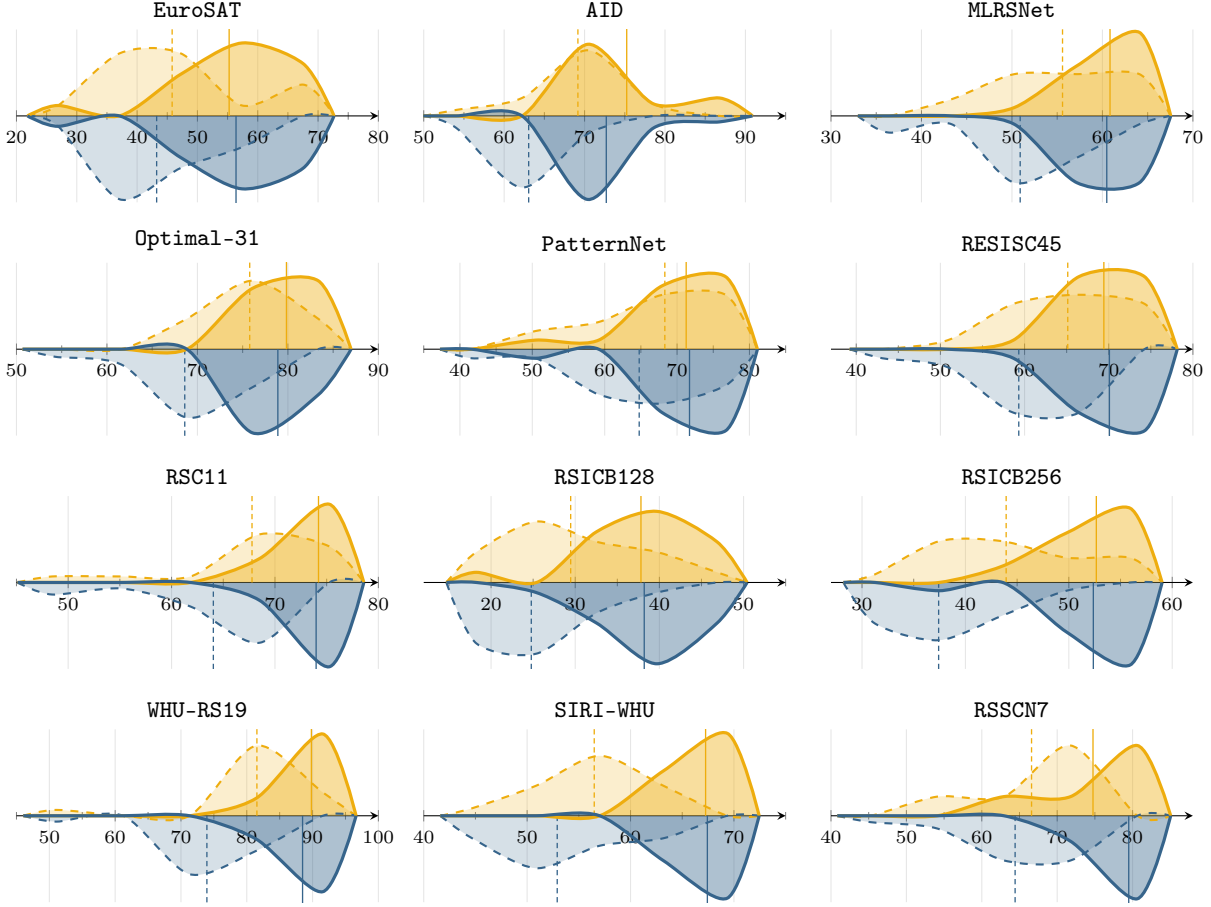

\noindent
\centering
\makebox[\textwidth][c]{%
\begin{tikzpicture}

\begin{groupplot}[
    group style={
        group size=3 by 4,
        horizontal sep=6mm,
        vertical sep=8mm
    },
    width=0.29\textwidth,
    height=0.10\textheight,
    every axis plot/.append style={draw=none},
    scale only axis,
    ymin=-1, ymax=1,
    axis x line=middle,
    axis y line=none,
    ytick=\empty,
    title style={font=\small, yshift=-5pt},
    xmajorgrids,
    xticklabel style={font=\scriptsize},
    grid style={gray!20},
]

\pgfmathsetmacro{\dx}{0}

\nextgroupplot[
    title={\eurosat},
    xmin=20, xmax=80,
    ymin=-0.24, ymax=0.24,
    minor x tick num=1,
    xtick distance=10,
]
\input{plots/violin_ret/eurosat.tex}

\nextgroupplot[
    title={\aid},
    xmin=50, xmax=95,
    ymin=-0.52, ymax=0.52,
    minor x tick num=1,
    xtick distance=10
]
\input{plots/violin_ret/aid.tex}

\nextgroupplot[
    title={\mlrsnet},
    xmin=30, xmax=70,
    ymin=-0.44, ymax=0.44,
    minor x tick num=1,
    xtick distance=10
]
\input{plots/violin_ret/mlrsnet.tex}

\nextgroupplot[
    title={\optimal},
    xmin=50, xmax=90,
    ymin=-0.46, ymax=0.46,
    minor x tick num=1,
    xtick distance=10
]
\input{plots/violin_ret/optimal31.tex}

\nextgroupplot[
    title={\patternnet},
    xmin=35, xmax=85,
    ymin=-0.32, ymax=0.32,
    minor x tick num=1,
    xtick distance=10
]
\input{plots/violin_ret/patternnet.tex}

\nextgroupplot[
    title={\resisc},
    xmin=37, xmax=80,
    ymin=-0.36, ymax=0.36,
    minor x tick num=1,
    xtick distance=10
]
\input{plots/violin_ret/resisc45.tex}

\nextgroupplot[
    title={\rsc},
    xmin=45, xmax=80,
    ymin=-0.63, ymax=0.63,
    minor x tick num=1,
    xtick distance=10
]
\input{plots/violin_ret/rsc11.tex}

\nextgroupplot[
    title={\rsicblow},
    xmin=12, xmax=55,
    ymin=-0.35, ymax=0.35,
    minor x tick num=1,
    xtick distance=10
]
\input{plots/violin_ret/rsicb128.tex}

\nextgroupplot[
    title={\rsicbhigh},
    xmin=27, xmax=62,
    ymin=-0.5, ymax=0.5,
    minor x tick num=1,
    xtick distance=10
]
\input{plots/violin_ret/rsicb256.tex}

\nextgroupplot[
    title={\whurs},
    xmin=45, xmax=100,
    ymin=-0.43, ymax=0.43,
    minor x tick num=1,
    xtick distance=10
]
\input{plots/violin_ret/whurs19.tex}

\nextgroupplot[
    title={\siriwhu},
    xmin=40, xmax=75,
    ymin=-0.55, ymax=0.55,
    minor x tick num=1,
    xtick distance=10
]
\input{plots/violin_ret/siriwhu.tex}

\nextgroupplot[
    title={\rsscn},
    xmin=40, xmax=88,
    ymin=-0.45, ymax=0.45,
    minor x tick num=1,
    xtick distance=10
]
\input{plots/violin_ret/rsscn7.tex}

\end{groupplot}

\end{tikzpicture}
}
\caption{Dataset-wise fixed-bin histograms of text-to-image retrieval mAP, across all evaluated models for \textcolor{\dstempcol}{\dst} and \textcolor{\llamacol}{\llama} generated descriptions. Dashed: results under the non-centered setting, solid: results under the text-side embedding centering.}
\label{fig:violin-ret}
\vspace{-15pt}
\end{figure*}
\paragraph{Performance distributions} Figures~\ref{fig:violin-cls} and~\ref{fig:violin-ret} visualize the distributions of downstream performance before and after query-side embedding centering for zero-shot image classification and text-to-image retrieval, respectively. Across both figures, the post-centering distributions are shifted to the right for the vast majority of datasets, indicating systematic improvements in both classification accuracy and retrieval mAP. Beyond the improvement in average performance, centering often shifts the entire distribution toward a higher-performing regime, with the post-centering curves in many cases becoming not only right-shifted but also more concentrated. This indicates that the effect of centering is broad and stabilizing, improving performance consistently across model and backbone instantiations rather than merely benefiting a small number of favorable cases. The improvement is generally more pronounced for \llama than for \dst, particularly in the retrieval setting, where the blue solid distributions frequently exhibit larger displacements than the yellow ones. At the same time, the magnitude of the shift remains dataset-dependent, with some benchmarks showing moderate gains and others exhibiting a much stronger redistribution toward higher scores.

% \textcolor{red}{For each dataset, we aggregate performance across VLM variants for \dst and \llama using mirrored smoothed histograms.}
% For each dataset, dashed curves denote the vanilla setting and solid curves the centered setting, with yellow corresponding to \dst and blue to \llama.

\input{plots/mpvr_voting}
\subsection{Ablation Study}
\paragraph{Top-k voting}
We further investigate an alternative strategy for zero-shot image classification based on top-$k$ voting. Instead of averaging all description embeddings of a class into a single prototype, we retain the individual text embeddings and, for each image, rank all candidate descriptions based on cosine similarity. The final class prediction is then obtained by majority voting among the classes associated with the top-k most similar text embeddings. Figure~\ref{fig:cls-voting} reports the average classification accuracy across datasets for different values of $k$, considering both \dst and \llama descriptions across the evaluated VLMs. The optimal value of $k$ is model-dependent, indicating that the contribution of this approach varies with the underlying geometry of each vision-language model. In the majority of cases, simple embedding averaging remains a strong upper reference, with top-$k$ voting yielding at most modest gains for the best selected $k$. Overall, the results indicate that although top-$k$ voting can occasionally yield marginal gains, it does not consistently outperform simple prototype averaging, whose stability and effectiveness make it the preferred default choice.
\newcommand{\imgpos}[2][0.11\textwidth]{% 
\tikz\node[ draw=green!70!black, line width=1.4pt, inner sep=1pt ]{\includegraphics[width=#1]{#2}};}
\newcommand{\imgneg}[2][0.11\textwidth]{% 
\tikz\node[ draw=red!70!black, line width=1.4pt, inner sep=1pt ]{\includegraphics[width=#1]{#2}};}

\section{Qualitative Results}
\begin{figure*}[t!]
\centering
\scriptsize
\setlength{\tabcolsep}{2pt}
\resizebox{\textwidth}{!}{%
\begin{tabular}{c c c c@{\hspace{3mm}} c c c @{\hspace{3mm}}c c c}
\toprule
& \multicolumn{3}{c}{\textbf{Chaparral}}
& \multicolumn{3}{c}{\textbf{Crossroads}}
& \multicolumn{3}{c}{\textbf{Field}} \\
\cmidrule(lr){2-4} \cmidrule(lr){5-7} \cmidrule(lr){8-10}
\textbf{Model}
& \textbf{R1} & \textbf{R2} & \textbf{R3}
& \textbf{R1} & \textbf{R2} & \textbf{R3}
& \textbf{R1} & \textbf{R2} & \textbf{R3} \\
\midrule

\dst
& \imgpos{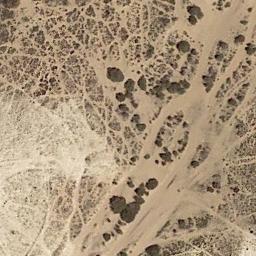}
& \imgneglabel{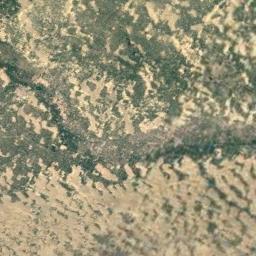}{Desert}
& \imgneglabel{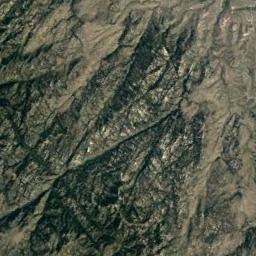}{Mountain}
& \imgneglabel{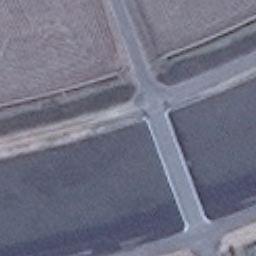}{Bridge}
& \imgneglabel{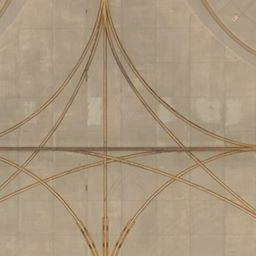}{Runway}
& \imgpos{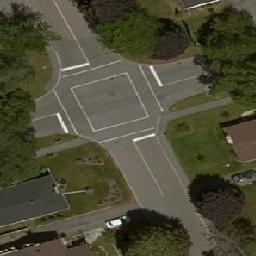}
& \imgpos{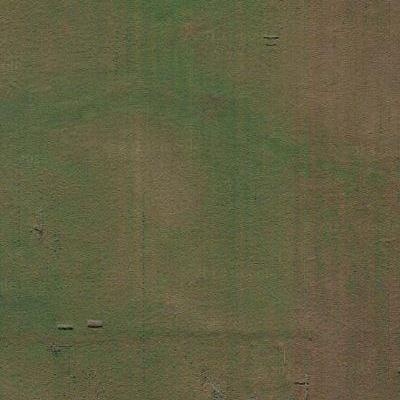}
& \imgneglabel{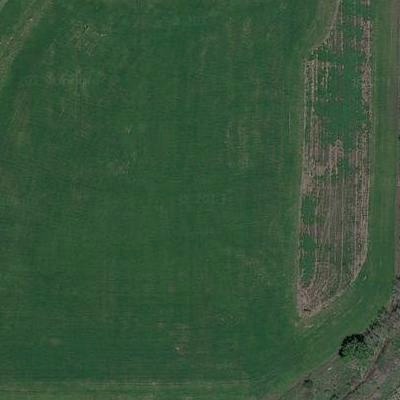}{Grass}
& \imgneglabel{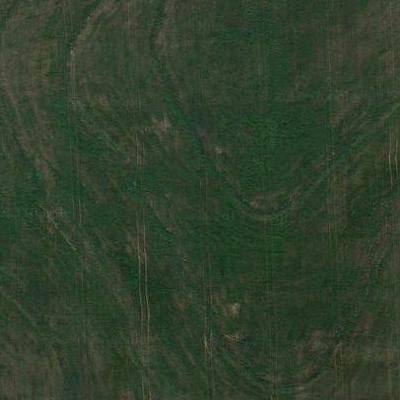}{Grass}
\\

\llama
& \imgneglabel{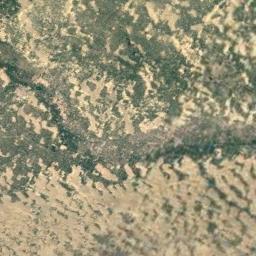}{Desert}
& \imgneglabel{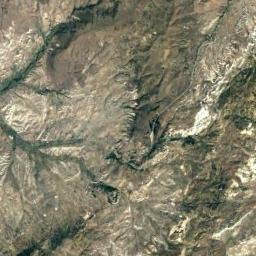}{Mountain}
& \imgneglabel{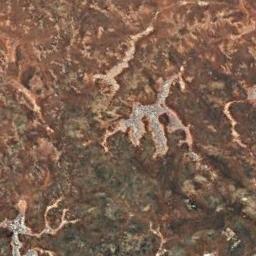}{Desert}
& \imgneglabel{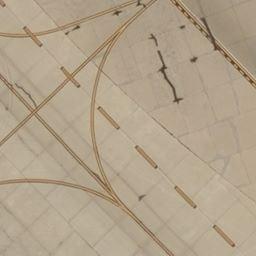}{Runway}
& \imgneglabel{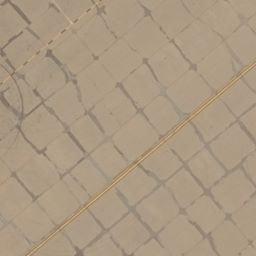}{Runway}
& \imgneglabel{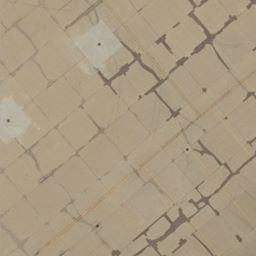}{Runway}
& \imgpos{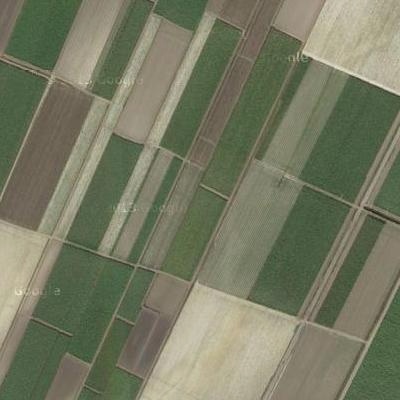}
& \imgpos{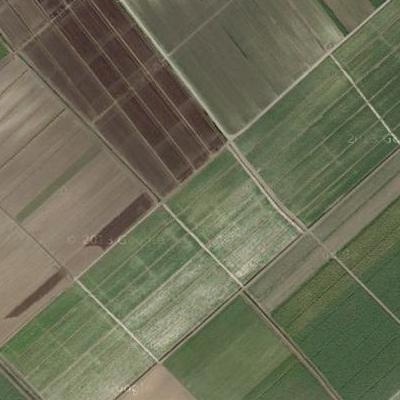}
& \imgpos{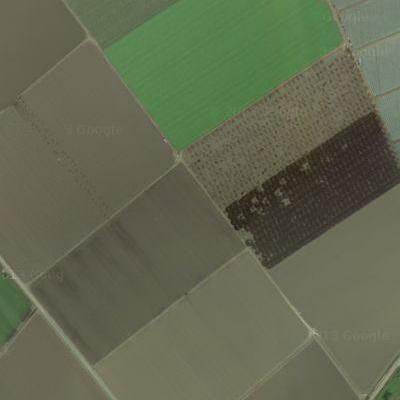}
\\

\llama\ Calibrated
& \imgpos{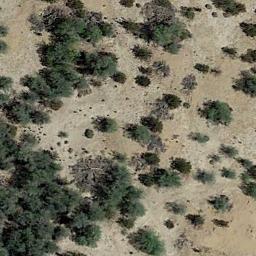}
& \imgpos{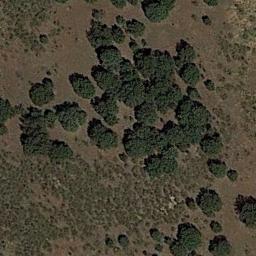}
& \imgpos{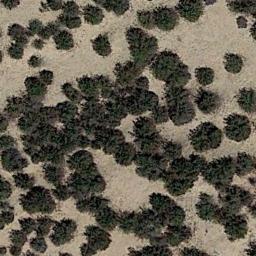}
& \imgpos{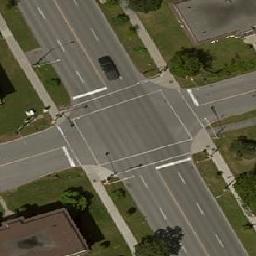}
& \imgpos{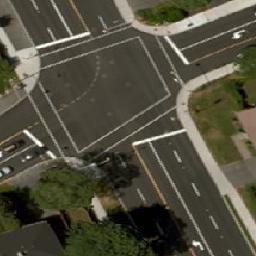}
& \imgpos{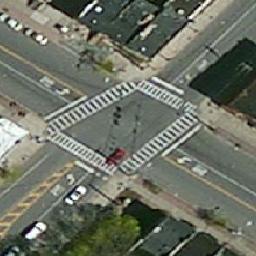}
& \imgpos{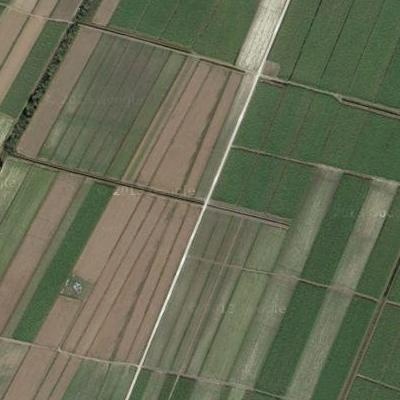}
& \imgpos{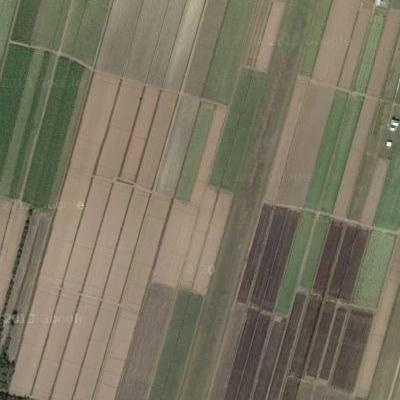}
& \imgpos{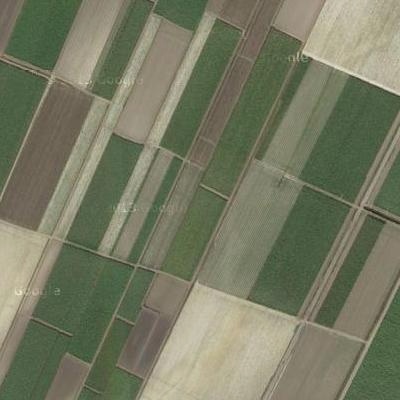}
\\

\bottomrule
\end{tabular}%
}

\caption{Qualitative text-to-image retrieval examples for the GeoRSCLIP ViT-L/14 vision-language model. For each class query, we show the top-3 retrieved images under three prompting variants: \dst, \llama, and \llama with calibration. The selected classes (\textit{Chaparral}, \textit{Crossroads}, and \textit{Field}) originate from different datasets. Green borders denote correct retrievals, whereas red borders indicate false positives.}

\label{fig:top3a}
\end{figure*}

\begin{figure*}[t!]
\centering
\scriptsize
\setlength{\tabcolsep}{2pt}
\resizebox{\textwidth}{!}{%
\begin{tabular}{c c c c@{\hspace{3mm}} c c c @{\hspace{3mm}}c c c}
\toprule
& \multicolumn{3}{c}{\textbf{Highway}}
& \multicolumn{3}{c}{\textbf{Permanent Crop Land}}
& \multicolumn{3}{c}{\textbf{Dense Residential}}\\
\cmidrule(lr){2-4} \cmidrule(lr){5-7} \cmidrule(lr){8-10}
\textbf{Model}
& \textbf{R1} & \textbf{R2} & \textbf{R3}
& \textbf{R1} & \textbf{R2} & \textbf{R3}
& \textbf{R1} & \textbf{R2} & \textbf{R3}\\
\midrule

\dst
& \imgpos{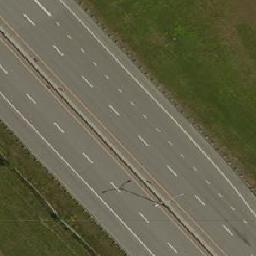}
& \imgpos{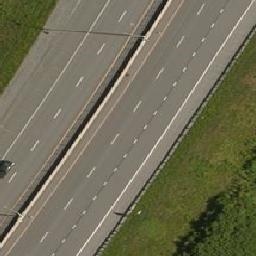}
& \imgneglabel{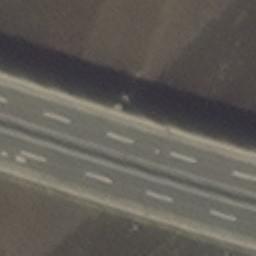}{Bridge}
& \imgneglabel{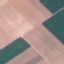}{Annual Crop}
& \imgneglabel{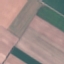}{Annual Crop}
& \imgneglabel{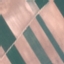}{Annual Crop}
&
\imgneglabel{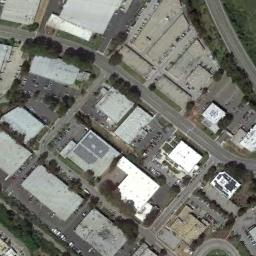}{Industrial}
& \imgneglabel{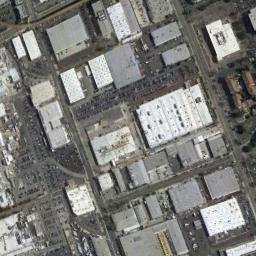}{Industrial}
& \imgneglabel{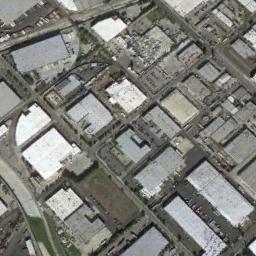}{Industrial}
\\

\llama
& \imgneglabel{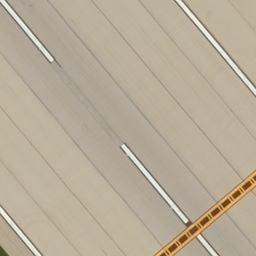}{Runway}
& \imgneglabel{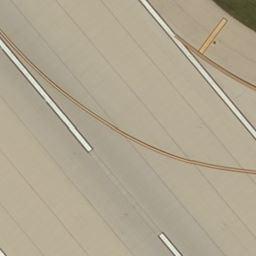}{Runway}
& \imgneglabel{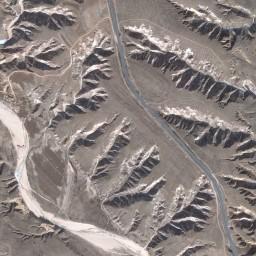}{Mountain}
& \imgpos{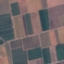}
& \imgneglabel{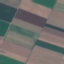}{Annual Crop}
& \imgpos{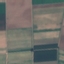}
&
\imgpos{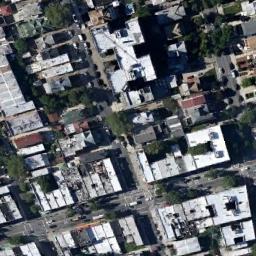}
& \imgpos{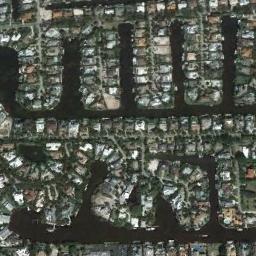}
& \imgpos{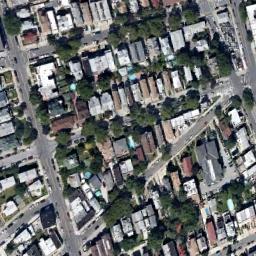}
\\

\llama\ Calibrated
& \imgpos{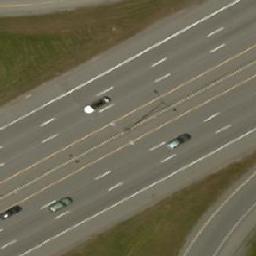}
& \imgpos{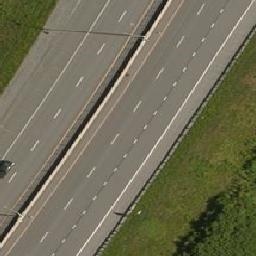}
& \imgpos{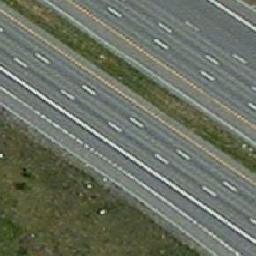}
& \imgpos{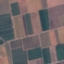}
& \imgpos{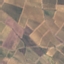}
& \imgpos{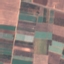}
&
\imgpos{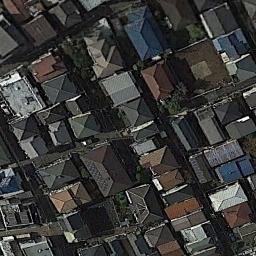}
& \imgpos{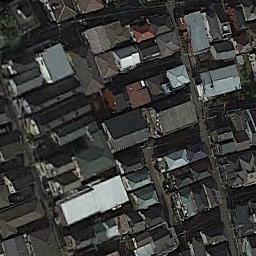}
& \imgpos{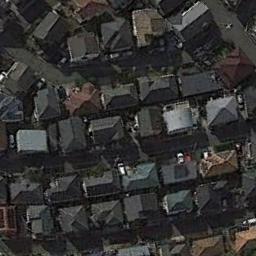}
\\

\bottomrule
\end{tabular}%
}

\caption{Additional qualitative text-to-image retrieval examples for the GeoRSCLIP ViT-L/14 vision-language model, showing the top-3 retrieved images for \dst, \llama, and \llama with calibration. The displayed classes (\textit{Highway}, \textit{Permanent Crop Land}, and \textit{Dense Residential}) are drawn from different datasets. Green borders indicate correct retrievals and red borders indicate incorrect ones.}
\label{fig:top3b}
\end{figure*}

\paragraph{Qualitative analysis of embedding calibration} Figures~\ref{fig:top3a} and~\ref{fig:top3b} present qualitative text-to-image retrieval examples for the GeoRSCLIP ViT-L/14 vision-language model by showing the top-3 retrieved images for three prompting variants, namely \dst, \llama, and \llama with calibration. A clear visual trend emerges across all illustrated classes: the calibrated \llama setup yields the most consistent retrieval behavior, returning correct top-3 matches throughout the presented examples. In contrast, both \dst\ and the uncalibrated \llama setup produce multiple false positives, as indicated by the red borders. Importantly, the classes shown in these figures are drawn from different datasets, suggesting that the improvement is not specific to a single benchmark but instead reflects a more general enhancement in the compatibility between text queries and image embeddings. Overall, these qualitative examples are in strong agreement with the quantitative retrieval results, showing that meta-prompt-generated descriptions become substantially more reliable and precise once calibration is applied.
% Fixed-size label settings
\setlength{\retrievallabelwidth}{14mm}
\setlength{\retrievallabelheight}{3.2mm}
\setlength{\retrievallabeltextwidth}{13mm} % slightly smaller than box width
% Label font settings
\setlength{\retrievallabelfontsize}{5pt}
\setlength{\retrievallabellineskip}{4.6pt}

%-----------------------------------------------------------

%-----------------------------------------------------------

\begin{figure*}[ht!]
\centering
\scriptsize
\setlength{\tabcolsep}{2pt}
\resizebox{\textwidth}{!}{%
\begin{tabular}{c c c c@{\hspace{4mm}}c c c}
\toprule
& \multicolumn{3}{c}{\textbf{Pasture Land}}
& \multicolumn{3}{c}{\textbf{Desert}} \\
\cmidrule(lr){2-4} \cmidrule(lr){5-7}
\textbf{Model}
& \textbf{R1} & \textbf{R2} & \textbf{R3}
& \textbf{R1} & \textbf{R2} & \textbf{R3} \\
\midrule

CLIP$_{\text{ViT-L}}$
& \imgpos{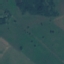}
& \imgneglabel{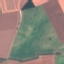}{Annual Crop}
& \imgneglabel{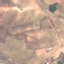}{Permanent Crop}
& \imgneglabel{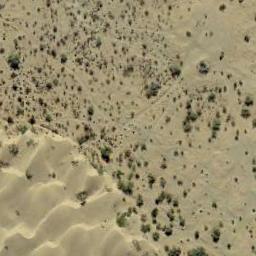}{Chaparral}
& \imgpos{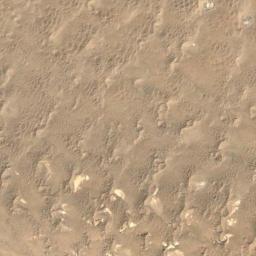}
& \imgpos{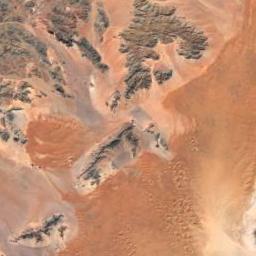}
\\

MetaCLIP-2.5B$_{\text{ViT-L}}$
& \imgpos{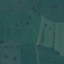}
& \imgpos{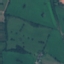}
& \imgneglabel{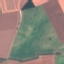}{Annual Crop}
& \imgpos{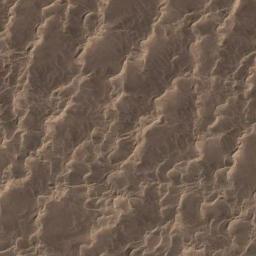}
& \imgpos{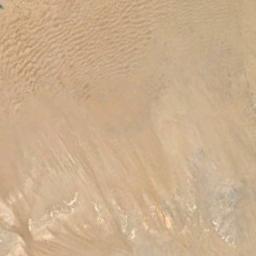}
& \imgpos{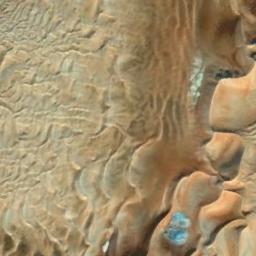}
\\

SigLIP2$_{\text{ViT-L}}$
& \imgpos{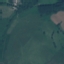}
& \imgneglabel{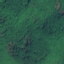}{Forest}
& \imgneglabel{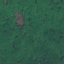}{Forest}
& \imgpos{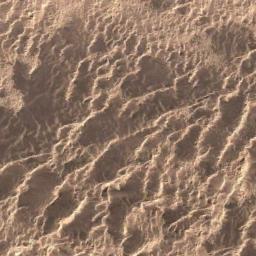}
& \imgpos{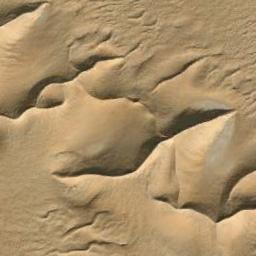}
& \imgneglabel{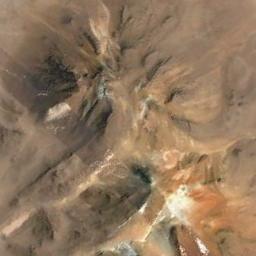}{Mountain}
\\

RemoteCLIP$_{\text{ViT-L}}$
& \imgpos{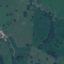}
& \imgneglabel{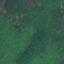}{Forest}
& \imgpos{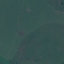}
& \imgneglabel{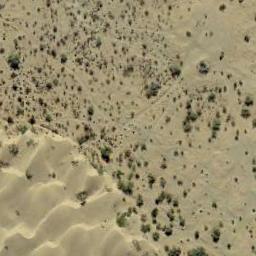}{Chaparral}
& \imgpos{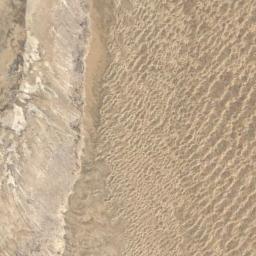}
& \imgneglabel{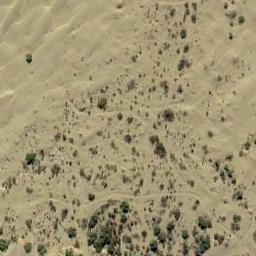}{Chaparral}
\\

GeoRSCLIP$_{\text{ViT-L}}$
& \imgpos{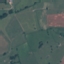}
& \imgpos{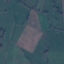}
& \imgpos{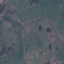}
& \imgpos{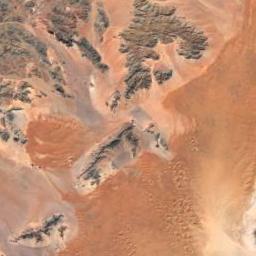}
& \imgpos{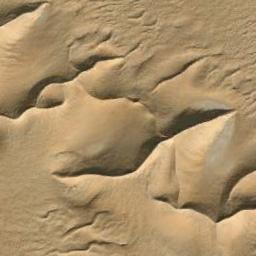}
& \imgneglabel{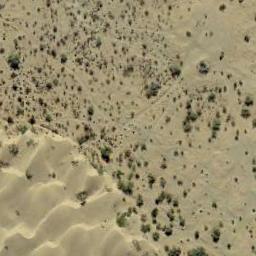}{Chaparral}
\\

SkyCLIP$_{\text{ViT-L}}$
& \imgpos{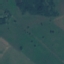}
& \imgneglabel{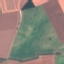}{Annual Crop}
& \imgpos{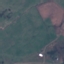}
& \imgpos{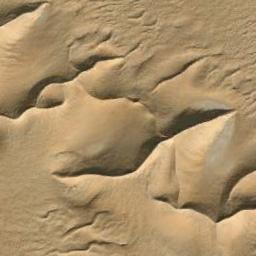}
& \imgpos{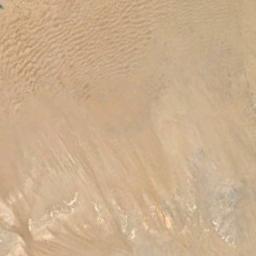}
& \imgpos{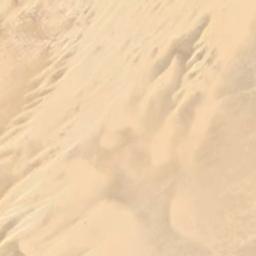}
\\

\bottomrule
\end{tabular}%
}

\caption{Qualitative comparison of text-to-image retrieval across different vision-language models under the \llama with calibration setup. For each class query, we show the top-3 retrieved images for CLIP ViT-L, MetaCLIP-2.5B ViT-L, SigLIP2 ViT-L, RemoteCLIP ViT-L, GeoRSCLIP ViT-L, and SkyCLIP ViT-L. The selected classes (\textit{Pasture Land} and \textit{Desert}) come from different datasets. Green borders denote correct retrievals, while red borders indicate incorrect ones.}
\label{fig:top3c}
\end{figure*}

\begin{figure*}[t!]
\centering
\scriptsize
\setlength{\tabcolsep}{2pt}
\resizebox{\textwidth}{!}{%
\begin{tabular}{c c c c@{\hspace{4mm}}c c c}
\toprule
& \multicolumn{3}{c}{\textbf{Mountain}}
& \multicolumn{3}{c}{\textbf{Airport}} \\
\cmidrule(lr){2-4} \cmidrule(lr){5-7}
\textbf{Model}
& \textbf{R1} & \textbf{R2} & \textbf{R3}
& \textbf{R1} & \textbf{R2} & \textbf{R3} \\
\midrule

CLIP$_{\text{ViT-L}}$
& \imgpos{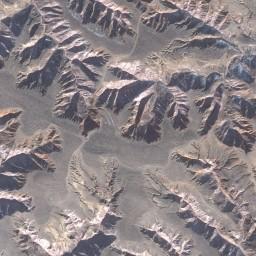}
& \imgpos{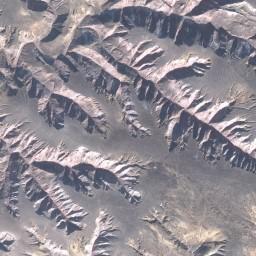}
& \imgneglabel{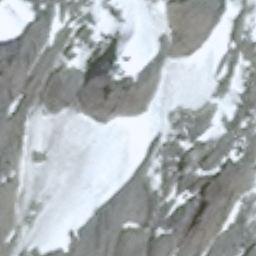}{Snow Mountain}
& \imgpos{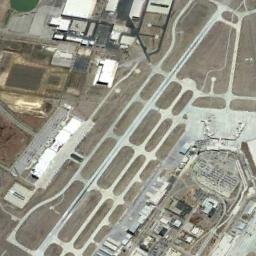}
& \imgpos{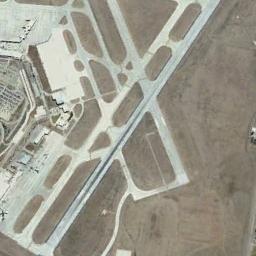}
& \imgpos{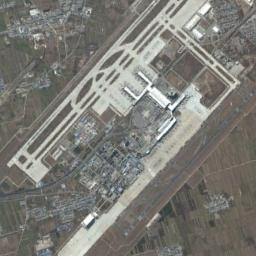}
\\

MetaCLIP-2.5B$_{\text{ViT-L}}$
& \imgpos{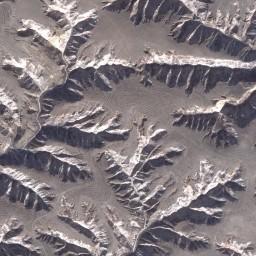}
& \imgpos{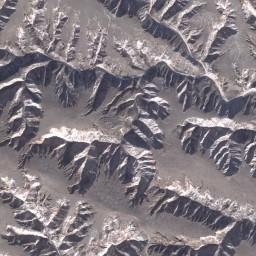}
& \imgpos{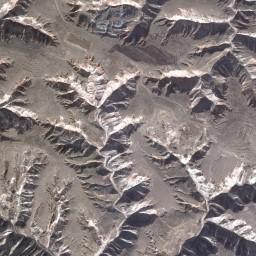}
& \imgpos{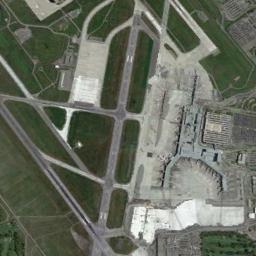}
& \imgpos{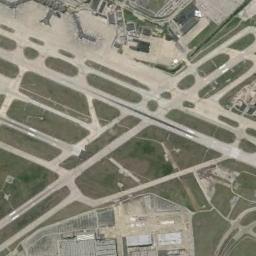}
& \imgneglabel{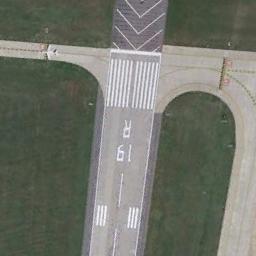}{Runway}
\\

SigLIP2$_{\text{ViT-L}}$
& \imgneglabel{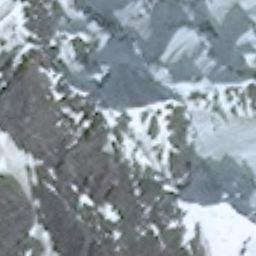}{Snow Mountain}
& \imgneglabel{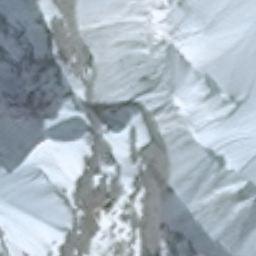}{Snow Mountain}
& \imgneglabel{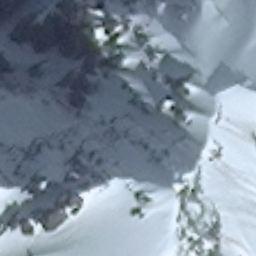}{Snow Mountain}
& \imgneglabel{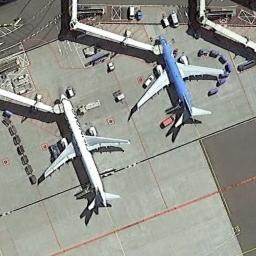}{Airplane}
& \imgneglabel{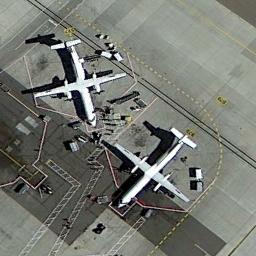}{Airplane}
& \imgneglabel{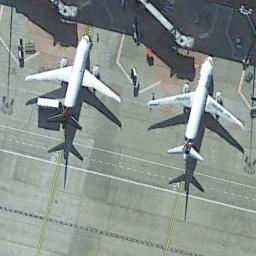}{Airplane}
\\

RemoteCLIP$_{\text{ViT-L}}$
& \imgpos{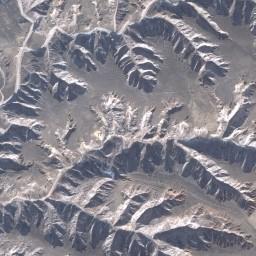}
& \imgpos{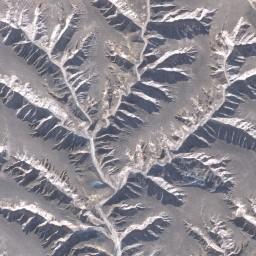}
& \imgpos{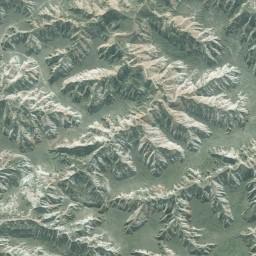}
& \imgpos{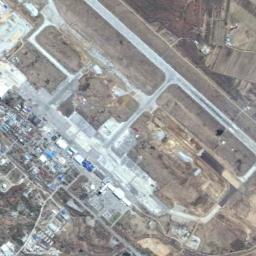}
& \imgpos{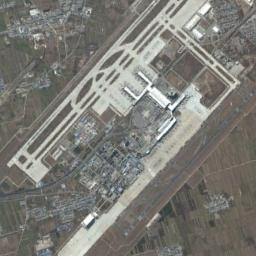}
& \imgpos{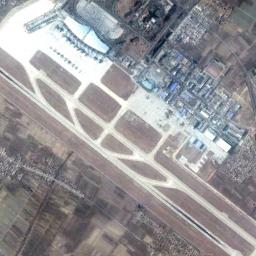}
\\

GeoRSCLIP$_{\text{ViT-L}}$
& \imgpos{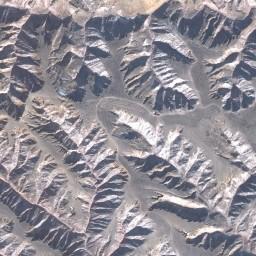}
& \imgpos{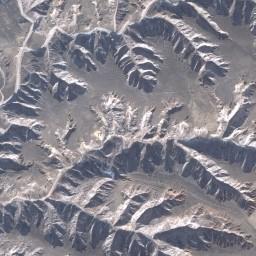}
& \imgpos{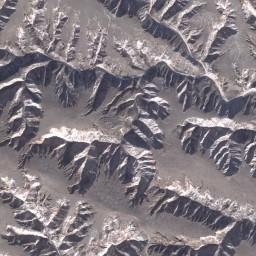}
& \imgpos{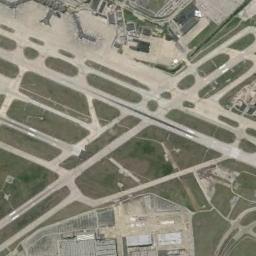}
& \imgpos{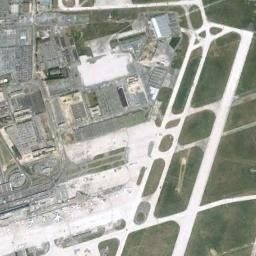}
& \imgpos{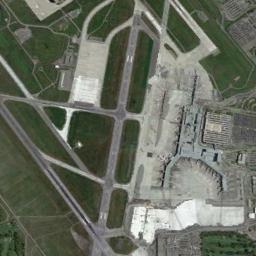}
\\

SkyCLIP$_{\text{ViT-L}}$
& \imgneglabel{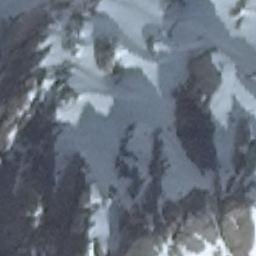}{Snow Mountain}
& \imgneglabel{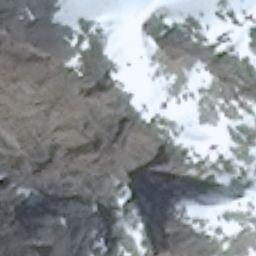}{Snow Mountain}
& \imgpos{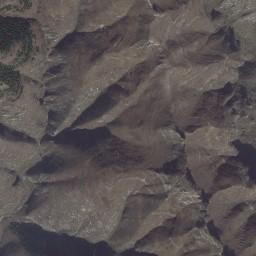}
& \imgpos{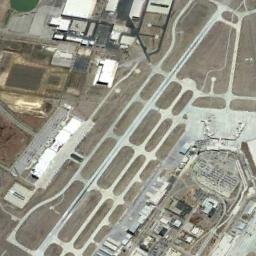}
& \imgpos{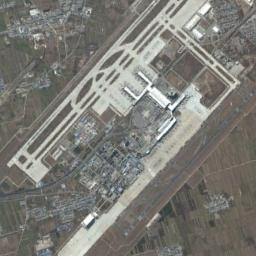}
& \imgpos{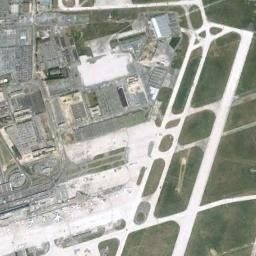}
\\

\bottomrule
\end{tabular}%
}
\caption{Additional qualitative comparison of text-to-image retrieval across different vision-language models using \llama with calibration. We report the top-3 retrieved images for CLIP ViT-L, MetaCLIP-2.5B ViT-L, SigLIP2 ViT-L, RemoteCLIP ViT-L, GeoRSCLIP ViT-L, and SkyCLIP ViT-L on the classes \textit{Mountain} and \textit{Airport}, which are drawn from different datasets. Green borders indicate correct retrievals and red borders indicate false positives.}
\label{fig:top3d}
\end{figure*}

\paragraph{Qualitative comparison across vision-language models} Figures~\ref{fig:top3c} and~\ref{fig:top3d} further examine qualitative text-to-image retrieval under the strongest prompting configuration, namely \llama with calibration, by comparing the top-3 retrieved images across different vision-language models. Specifically, the figures report results for CLIP ViT-L, MetaCLIP-2.5B ViT-L, SigLIP2 ViT-L, RemoteCLIP ViT-L, GeoRSCLIP ViT-L, and SkyCLIP ViT-L on classes drawn from different datasets. Even under this shared and highly consistent retrieval setup, clear differences remain across models. In particular, the remote-sensing-adapted backbones exhibit more stable and semantically precise retrieval behavior overall, with GeoRSCLIP ViT-L and SkyCLIP ViT-L emerging as the most consistent performers across the illustrated examples. By contrast, the general-purpose models show a larger number of residual false positives or less reliable ranking of the top retrieved images. These examples therefore complement the quantitative results by showing that, once calibration reduces prompt-side ambiguity, the quality of retrieval is increasingly governed by the underlying visual-language representation itself.
 
% Fixed-size label settings
\setlength{\retrievallabelwidth}{14mm}
\setlength{\retrievallabelheight}{3.2mm}
\setlength{\retrievallabeltextwidth}{13mm} % slightly smaller than box width
% Label font settings
\setlength{\retrievallabelfontsize}{5pt}
\setlength{\retrievallabellineskip}{4.6pt}

%-----------------------------------------------------------

%-----------------------------------------------------------
\section{Discussion}
Our study shows that semantic richness, by itself, is not sufficient to guarantee stronger zero-shot performance. In the remote sensing setting, richer LLM-generated descriptions can expand the semantic coverage of a class, but they also introduce variability that is not always well aligned with the frozen vision-language shared embedding space. This suggests that the benefit of richer language supervision is closely tied to the underlying learning regime. An important direction for future work is therefore to examine how semantic richness behaves beyond the strictly zero-shot setting, for example in few-shot scenarios or under trainable adaptation strategies that explicitly align longer and more expressive textual descriptions with image embeddings in the shared representation space.
\section{Conclusion}
In this work, we revisit zero-shot remote sensing through a systematic study of meta-prompting, prototype construction, and embedding calibration across 12 remote sensing benchmarks, multiple vision-language model families, and 5 open-source LLMs. Our results show that, although LLM-generated descriptions provide richer semantic content, they do not consistently improve downstream performance and often introduce additional variability in the text embedding space, whereas simple domain-specific templates remain strong and robust baselines. At the same time, we demonstrate that a significant part of this limitation is not only due to the generated descriptions themselves, but also to how they interact with the shared embedding space: query-side centering yields broad and consistent gains, with image-side centering improving zero-shot classification and text-side centering producing even stronger benefits for text-to-image retrieval. Across both tasks, remote-sensing-adapted VLMs consistently outperform general-purpose alternatives, confirming the value of domain-specialized representations for satellite imagery. In summary, our findings highlight a trade-off between semantic richness and robustness and show that simple embedding centering constitutes a highly effective mechanism for improving zero-shot remote sensing performance.
\section{Acknowledgments}
This work was supported by the Edge SpAIce Horizon Europe project under Grant 101135358.
We gratefully acknowledge the EuroHPC Joint Undertaking for awarding this project access to the EuroHPC supercomputer LEONARDO, hosted by CINECA (Italy) and the LEONARDO consortium through an EuroHPC Development Access call.

\section{Declaration of generative AI and AI-assisted technologies in the manuscript preparation process}
During the preparation of this work the authors used ChatGPT specifically to aid in the polishing and refinement of the text. LLMs were \emph{not} used for ideation, technical contributions, experimental design, data analysis, or related work retrieval. All research ideas, methodology, experiments, and conclusions presented are solely the work of the authors. After using this tool/service, the authors reviewed and edited the content as needed and take full responsibility for the content of the published article.

% --------------------------------------------------
% Appendix
% --------------------------------------------------
\clearpage
\appendix
\renewcommand{\thesection}{Appendix~\Alph{section}}
%\section*{Appendix}
\section{Dataset Details}
\label{app:datasets}
Table \ref{tab:datasets_overview} provides an overview of the 12 remote sensing scene classification datasets considered in this work, including the number of classes, number of images, image resolution, and when available, the corresponding ground sampling distance (GSD). Together, these benchmarks cover a wide spectrum of remote-sensing scenarios, with substantial variation in dataset size, spatial scale, and semantic complexity. This heterogeneity is particularly relevant for our evaluation setting, as it allows us to examine the robustness of zero-shot prompting and retrieval behavior across benchmarks with different visual characteristics and label-space structures.
\begin{table}[ht!]
\centering
\caption{Overview of remote sensing scene classification datasets used in our experiments.}
\label{tab:datasets_overview}
\footnotesize
\begin{tabular}{l|c|c|c|c}
\toprule
\textbf{Dataset} &
\textbf{\# Classes} &
\textbf{\# Images} &
\textbf{Resolution} &
\textbf{GSD (m)} \\
\midrule
\rsscn \cite{zou2015deep}          & 7  & 2{,}800   & $400\times400$ & --          \\
\eurosat \cite{{helber2019eurosat}}     & 10 & 27{,}000  & $64\times64$   & 10          \\
\rsc \cite{zhao2016feature}           & 11 & 1{,}232   & $512\times512$ & 0.2         \\
\siriwhu \cite{zhao2015dirichlet}      & 12 & 2{,}400   & $200\times200$ & 2           \\
\whurs \cite{xia2010structural}      & 19 & 1{,}005   & $600\times600$ & $\leq$~0.5  \\
\aid \cite{xia2017aid}                 & 30 & 10{,}000  & $600\times600$ & 0.5--8      \\
\optimal \cite{wang2018scene}        & 31 & 1{,}860   & $256\times256$ & --          \\
\rsicbhigh \cite{li2020rsi}              & 35 & 24{,}747  & $256\times256$ & 0.3--3      \\
\patternnet \cite{zhou2018patternnet}   & 38 & 30{,}400  & $256\times256$ & 0.06--4.7   \\
\resisc \cite{cheng2017remote}        & 45 & 31{,}500  & $256\times256$ & 0.2--30     \\
\rsicblow \cite{li2020rsi}              & 45 & 36{,}707  & $128\times128$ & 0.3--3      \\
\mlrsnet \cite{qi2020mlrsnet}           & 46 & 109{,}161 & $256\times256$ & 0.1--10     \\
\bottomrule
\end{tabular}
\end{table}

\section{Extended Results}
\label{app:extended-results}

\begin{table*}[ht!]
\centering
\caption{Zero-shot classification accuracy (\%) on all evaluated remote sensing datasets under the Simple Template (\st) baseline. Results are reported across the full set of vision-language models and backbone architectures.}
\vspace{-2pt}
\label{tab:cls_st}
\resizebox{\textwidth}{!}{
\begin{tabular}{lcccccccccccccc}
\toprule
\SC{Model} & \SC{Arch}
& \rotatebox{90}{\eurosat}
& \rotatebox{90}{\aid}
& \rotatebox{90}{\mlrsnet}
& \rotatebox{90}{\optimal}
& \rotatebox{90}{\patternnet}
& \rotatebox{90}{\resisc}
& \rotatebox{90}{\rsc}
& \rotatebox{90}{\rsicblow}
& \rotatebox{90}{\rsicbhigh}
& \rotatebox{90}{\whurs}
& \rotatebox{90}{\siriwhu}
& \rotatebox{90}{\rsscn}
& \rotatebox{90}{\SC{Avg}} \\
\midrule
\multirow{2}{*}{CLIP\cite{clip}} & ViT-B/16 & 44.1 & 66.6 & 54.4 & 70.1 & 62.9 & 62.1 & 62.9 & 26.9 & 38.4 & 82.8 & 52.2 & 64.3 & 57.3 \\
& ViT-L/14 & 63.0 & 69.5 & 63.9 & 76.1 & 71.8 & 67.0 & 67.1 & 38.0 & 47.6 & 85.1 & 58.8 & 61.5 & 64.1 \\
\midrule
\multirow{2}{*}{MetaCLIP-400M\cite{xu2023demystifying}} & ViT-B/16 & 55.2 & 71.9 & 57.0 & 76.5 & 68.2 & 66.2 & 65.9 & 31.0 & 41.0 & 89.0 & 59.7 & 66.6 & 62.3 \\
& ViT-L/14 & 52.6 & 69.3 & 60.1 & 80.0 & 74.5 & 70.0 & 55.2 & 33.4 & 45.3 & 88.8 & 65.9 & 64.5 & 63.3 \\
\midrule
\multirow{2}{*}{MetaCLIP-2.5B\cite{xu2023demystifying}} & ViT-B/16 & 46.6 & 71.0 & 62.5 & 76.0 & 70.0 & 68.5 & 61.7 & 31.1 & 40.5 & 88.2 & 62.9 & 63.1 & 61.9 \\
& ViT-L/14 & 55.4 & 78.3 & \underline{70.1} & 83.1 & \textbf{83.5} & 74.1 & 67.0 & 39.1 & 41.9 & 91.0 & 64.3 & 65.7 & 67.8 \\
\midrule
\multirow{2}{*}{SigLIP\cite{siglip}} & ViT-B/16 & 45.0 & 70.9 & 57.3 & 74.5 & 63.8 & 64.4 & 64.5 & 24.2 & 38.1 & 84.4 & 62.2 & 65.6 & 59.6 \\
& ViT-L/16 & 53.3 & 73.5 & 60.2 & 81.3 & 67.6 & 68.5 & 69.1 & 29.4 & 42.5 & 88.9 & 62.3 & 62.7 & 63.3 \\
\midrule
\multirow{2}{*}{SigLIP2 \cite{siglip2}} & ViT-B/16 & 46.4 & 75.3 & 63.5 & 82.6 & 71.6 & 71.1 & 61.9 & 30.3 & 43.2 & 88.8 & 61.8 & 65.6 & 63.5 \\
& ViT-L/16 & 54.1 & 75.3 & \textbf{73.3} & \textbf{87.1} & 78.8 & \textbf{77.8} & \textbf{77.8} & 37.3 & 47.4 & 92.0 & 66.1 & 70.7 & 69.8 \\
\midrule
\multirow{3}{*}{RemoteCLIP
\cite{liu2023remoteclip}} & RN-50 & 20.5 & \underline{87.2} & 41.7 & 63.0 & 44.1 & 49.7 & 63.8 & 16.3 & 33.2 & 93.6 & 55.7 & 48.3 & 51.4 \\
& ViT-B/32 & 38.4 & \textbf{89.3} & 55.3 & 76.2 & 58.5 & 66.3 & 60.3 & 27.1 & 42.0 & \underline{93.7} & 60.8 & 58.8 & 60.6 \\
& ViT-L/14 & 43.1 & 82.3 & 62.0 & 82.3 & 64.1 & \underline{75.1} & 62.6 & 34.4 & 48.9 & 86.7 & \textbf{67.8} & 63.2 & 64.4 \\
\midrule
\multirow{2}{*}{GeoRSCLIP\cite{zhang2024rs5m}}  & ViT-B/32 & 48.6 & 70.9 & 65.3 & 79.6 & 75.9 & 70.1 & 65.9 & 26.9 & 45.9 & 90.0 & 56.2 & \textbf{76.1} & 64.3 \\
& ViT-L/14 & 67.3 & 76.6 & 68.7 & \underline{86.1} & \underline{81.5} & 74.8 & \underline{75.8} & 39.8 & \textbf{54.8} & \textbf{94.3} & 62.0 & \underline{72.9} & \textbf{71.2} \\
\midrule
CLIP LAION-RS \cite{wang2024skyscript} & ViT-L/14 & \textbf{70.4} & 72.8 & 66.5 & 80.5 & 76.7 & 72.0 & 70.9 & \underline{41.7} & 50.6 & 89.7 & 64.2 & 66.5 & 68.5 \\
\midrule
SkyCLIP \cite{wang2024skyscript} & ViT-L/14 & \underline{70.0} & 74.0 & 69.1 & 82.3 & 79.8 & 72.0 & 73.1 & \textbf{49.2} & \underline{53.4} & 90.0 & \underline{66.2} & 68.8 & \underline{70.7} \\
\bottomrule
\end{tabular}
}
\vspace{-5pt}
\end{table*}
\begin{table*}[ht!]
\centering
\caption{Zero-shot classification accuracy (\%) on all evaluated remote sensing datasets under the meta-prompting approach with \deepseek. Results are reported across the full set of vision-language models and backbone architectures.}
\label{tab:cls_deepseek}
\vspace{-5pt}
\resizebox{\textwidth}{!}{
\begin{tabular}{lcccccccccccccc}
\toprule
\SC{Model} & \SC{Arch}
& \rotatebox{90}{\eurosat}
& \rotatebox{90}{\aid}
& \rotatebox{90}{\mlrsnet}
& \rotatebox{90}{\optimal}
& \rotatebox{90}{\patternnet}
& \rotatebox{90}{\resisc}
& \rotatebox{90}{\rsc}
& \rotatebox{90}{\rsicblow}
& \rotatebox{90}{\rsicbhigh}
& \rotatebox{90}{\whurs}
& \rotatebox{90}{\siriwhu}
& \rotatebox{90}{\rsscn}
& \rotatebox{90}{\SC{Avg}} \\
\midrule
\multirow{2}{*}{CLIP} & ViT-B/16 & 52.8 & 67.1 & 56.6 & 72.9 & 65.1 & 64.9 & 62.5 & 28.6 & 37.7 & 85.2 & 56.6 & 66.2 & 59.7 \\
& ViT-L/14 & 64.5 & 70.9 & 64.0 & 78.0 & 73.3 & 70.1 & 62.0 & 32.3 & 43.7 & 87.4 & 61.8 & 63.0 & 64.3 \\
\midrule
\multirow{2}{*}{MetaCLIP-400M} & ViT-B/16 & 55.5 & 67.0 & 60.3 & 75.8 & 69.7 & 67.6 & 67.9 & 31.4 & 40.5 & 87.5 & 57.1 & 70.9 & 62.6 \\
& ViT-L/14 & 53.0 & 70.8 & 63.2 & 76.7 & 75.7 & 71.4 & 61.1 & 30.5 & 40.4 & 88.7 & 64.2 & 69.9 & 63.8 \\
\midrule
\multirow{2}{*}{MetaCLIP-2.5B} & ViT-B/16 & 49.3 & 67.9 & 63.9 & 75.2 & 71.8 & 68.4 & 59.5 & 33.0 & 38.1 & 84.5 & 60.7 & 68.6 & 61.7 \\
& ViT-L/14 & 63.0 & 73.0 & 67.1 & 80.8 & 79.5 & 72.8 & 63.2 & 39.5 & 41.9 & 87.5 & 62.6 & 68.3 & 66.6 \\
\midrule
\multirow{2}{*}{SigLIP} & ViT-B/16 & 44.6 & 67.9 & 58.7 & 73.6 & 63.7 & 66.4 & 62.1 & 25.9 & 37.8 & 80.9 & 58.1 & 65.9 & 58.8 \\
& ViT-L/16 & 53.6 & 73.7 & 64.5 & 79.3 & 68.8 & 73.1 & 70.6 & 30.3 & 42.8 & 88.0 & 60.2 & 69.5 & 64.5 \\
\midrule
\multirow{2}{*}{SigLIP2} & ViT-B/16 & 46.8 & 72.0 & 63.6 & 78.4 & 67.8 & 70.6 & 64.9 & 30.2 & 42.1 & 85.9 & 58.5 & 65.6 & 62.2 \\
& ViT-L/16 & 58.4 & 74.4 & 68.4 & 79.7 & 74.3 & \underline{76.0} & 68.1 & 34.4 & 43.4 & 90.4 & 61.8 & 71.3 & 66.7 \\
\midrule
\multirow{3}{*}{RemoteCLIP} & RN-50 & 25.2 & \underline{83.8} & 46.7 & 70.2 & 46.5 & 59.0 & 63.3 & 23.2 & 32.7 & \underline{93.6} & 64.3 & 65.9 & 56.2 \\
& ViT-B/32 & 38.7 & \textbf{87.8} & 58.5 & 79.3 & 60.9 & 71.6 & 67.5 & 26.2 & 38.8 & \textbf{96.0} & \textbf{71.5} & 68.9 & 63.8 \\
& ViT-L/14 & 43.1 & 81.2 & 61.4 & 81.5 & 63.3 & 73.7 & 64.7 & 30.3 & 47.2 & 91.5 & \underline{70.6} & 67.1 & 64.6 \\
\midrule
\multirow{2}{*}{GeoRSCLIP} & ViT-B/32 & 55.1 & 72.7 & 66.1 & 79.3 & 76.0 & 70.3 & \underline{72.6} & 32.3 & 49.7 & 88.9 & 60.3 & \textbf{78.6} & 66.8 \\
& ViT-L/14 & \underline{69.2} & 75.8 & \textbf{70.6} & \textbf{83.8} & \underline{80.9} & \textbf{76.1} & \textbf{74.2} & \underline{40.3} & \underline{53.0} & 90.4 & 61.6 & \underline{74.3} & \textbf{70.9} \\
\midrule
CLIP LAION-RS & ViT-L/14 & 68.6 & 72.4 & 65.0 & 78.9 & 75.5 & 71.3 & 66.0 & 37.4 & 48.4 & 89.7 & 62.5 & 68.6 & 67.0 \\
\midrule
SkyCLIP & ViT-L/14 & \textbf{72.2} & 74.6 & \underline{69.2} & \underline{81.9} & \textbf{80.9} & 74.1 & 70.5 & \textbf{45.3} & \textbf{55.1} & 90.8 & 63.3 & 69.7 & \underline{70.6} \\
\bottomrule
\end{tabular}
}
\vspace{-5pt}
\end{table*}
\begin{table*}[ht!]
\centering
\caption{Zero-shot classification accuracy (\%) on all evaluated remote sensing datasets under the meta-prompting approach with \qwen. Results are reported across the full set of vision-language models and backbone architectures.}
\label{tab:cls_qwen}
\vspace{-5pt}
\resizebox{\textwidth}{!}{
\begin{tabular}{lcccccccccccccc}
\toprule
\SC{Model} & \SC{Arch}
& \rotatebox{90}{\eurosat}
& \rotatebox{90}{\aid}
& \rotatebox{90}{\mlrsnet}
& \rotatebox{90}{\optimal}
& \rotatebox{90}{\patternnet}
& \rotatebox{90}{\resisc}
& \rotatebox{90}{\rsc}
& \rotatebox{90}{\rsicblow}
& \rotatebox{90}{\rsicbhigh}
& \rotatebox{90}{\whurs}
& \rotatebox{90}{\siriwhu}
& \rotatebox{90}{\rsscn}
& \rotatebox{90}{\SC{Avg}}  \\
\midrule
\multirow{2}{*}{CLIP} & ViT-B/16 & 53.0 & 64.7 & 55.9 & 72.7 & 63.9 & 63.9 & 58.6 & 31.0 & 39.5 & 84.3 & 53.0 & 70.7 & 59.3 \\
& ViT-L/14 & 62.5 & 67.8 & 63.8 & 76.3 & 72.9 & 69.5 & 60.7 & 37.4 & 49.0 & 86.6 & 58.0 & 70.1 & 64.5 \\
\midrule
\multirow{2}{*}{MetaCLIP-400M} & ViT-B/16 & 58.5 & 64.0 & 58.2 & 76.9 & 67.8 & 66.8 & 63.6 & 34.0 & 43.0 & 89.2 & 54.1 & 72.3 & 62.4 \\
& ViT-L/14 & 56.4 & 69.4 & 62.5 & 76.3 & 73.5 & 70.3 & 63.2 & 33.8 & 44.1 & 88.1 & 62.1 & 75.2 & 64.6 \\
\midrule
\multirow{2}{*}{MetaCLIP-2.5B} & ViT-B/16 & 50.6 & 66.7 & 63.2 & 77.4 & 70.0 & 69.2 & 54.8 & 37.2 & 44.6 & 89.6 & 57.7 & 72.8 & 62.8 \\
& ViT-L/14 & 65.0 & 70.4 & 67.2 & 80.4 & 78.7 & 71.2 & 55.8 & 41.7 & 44.9 & 88.3 & 60.8 & 72.2 & 66.4 \\
\midrule
\multirow{2}{*}{SigLIP} & ViT-B/16 & 44.3 & 65.2 & 57.9 & 74.1 & 65.0 & 65.3 & 62.3 & 26.2 & 39.4 & 79.5 & 51.1 & 68.4 & 58.2 \\
& ViT-L/16 & 54.6 & 70.9 & 64.5 & 78.8 & 70.9 & 71.9 & 69.0 & 31.6 & 47.5 & 87.7 & 59.4 & 73.3 & 65.0 \\
\midrule
\multirow{2}{*}{SigLIP2} & ViT-B/16 & 49.3 & 69.9 & 62.6 & 78.2 & 69.3 & 70.2 & 61.3 & 32.8 & 46.2 & 86.5 & 56.1 & 67.3 & 62.5 \\
& ViT-L/16 & 59.7 & 71.9 & 67.4 & 81.2 & 76.8 & \textbf{74.7} & 67.9 & 35.7 & 47.8 & 90.1 & 59.2 & 70.9 & 66.9 \\
\midrule
\multirow{3}{*}{RemoteCLIP} & RN-50 & 26.2 & 73.9 & 45.8 & 69.7 & 46.4 & 56.8 & 62.3 & 22.8 & 33.6 & 89.8 & 59.6 & 66.8 & 54.5 \\
& ViT-B/32 & 38.3 & \textbf{79.9} & 58.5 & 79.0 & 59.1 & 71.6 & 67.7 & 28.1 & 38.3 & \textbf{94.6} & \underline{65.5} & 72.0 & 62.7 \\
& ViT-L/14 & 47.8 & \underline{76.5} & 61.5 & 81.9 & 63.2 & 74.1 & \textbf{70.2} & 30.5 & 49.2 & \underline{91.2} & \textbf{72.5} & 69.4 & 65.7 \\
\midrule
\multirow{2}{*}{GeoRSCLIP} & ViT-B/32 & 51.2 & 68.8 & 65.3 & 79.9 & 77.2 & 69.8 & 66.8 & 35.1 & 47.6 & 89.8 & 60.2 & \textbf{79.0} & 65.9 \\
& ViT-L/14 & 68.4 & 70.8 & \textbf{69.4} & \textbf{84.7} & \underline{79.8} & \underline{74.3} & \underline{69.1} & \underline{44.1} & \underline{55.5} & 90.3 & 59.0 & \underline{78.1} & \underline{70.3} \\
\midrule
CLIP LAION-RS & ViT-L/14 & \underline{73.0} & 68.6 & 64.6 & 78.5 & 74.9 & 71.1 & 63.9 & 38.3 & 52.5 & 89.7 & 61.6 & 75.3 & 67.7 \\
\midrule
SkyCLIP & ViT-L/14 & \textbf{75.7} & 70.4 & \underline{69.0} & \underline{82.0} & \textbf{80.5} & 73.4 & 68.3 & \textbf{47.2} & \textbf{56.0} & 91.1 & 61.3 & 74.7 & \textbf{70.8} \\
\bottomrule
\end{tabular}
}
\vspace{-5pt}
\end{table*}
\begin{table*}[ht!]
\centering
\caption{Zero-shot classification accuracy (\%) on all evaluated remote sensing datasets under the meta-prompting approach with \mixtral. Results are reported across the full set of vision-language models and backbone architectures.}
\label{tab:cls_mixtral}
\vspace{-5pt}
\resizebox{\textwidth}{!}{
\begin{tabular}{lcccccccccccccc}
\toprule
\SC{Model} & \SC{Arch}
& \rotatebox{90}{\eurosat}
& \rotatebox{90}{\aid}
& \rotatebox{90}{\mlrsnet}
& \rotatebox{90}{\optimal}
& \rotatebox{90}{\patternnet}
& \rotatebox{90}{\resisc}
& \rotatebox{90}{\rsc}
& \rotatebox{90}{\rsicblow}
& \rotatebox{90}{\rsicbhigh}
& \rotatebox{90}{\whurs}
& \rotatebox{90}{\siriwhu}
& \rotatebox{90}{\rsscn}
& \rotatebox{90}{\SC{Avg}} \\ 
\midrule
\multirow{2}{*}{CLIP} & ViT-B/16 & 53.9 & 68.5 & 56.7 & 73.9 & 66.8 & 63.7 & 62.1 & 26.3 & 35.8 & 84.7 & 54.5 & 64.4 & 59.3 \\
& ViT-L/14 & 65.5 & 72.2 & 65.0 & 80.6 & 75.0 & 68.8 & 65.6 & 34.1 & 46.2 & 87.5 & 59.7 & 64.0 & 65.4 \\
\midrule
\multirow{2}{*}{MetaCLIP-400M} & ViT-B/16 & 55.9 & 68.6 & 58.0 & 76.8 & 69.7 & 64.0 & 68.3 & 28.3 & 36.0 & 85.0 & 57.0 & 64.3 & 61.0 \\
& ViT-L/14 & 59.2 & 70.9 & 63.8 & 79.2 & 75.6 & 70.3 & 64.3 & 30.4 & 39.7 & 88.4 & 62.6 & 68.8 & 64.4 \\
\midrule
\multirow{2}{*}{MetaCLIP-2.5B} & ViT-B/16 & 52.1 & 70.3 & 61.2 & 75.3 & 71.8 & 66.4 & 62.8 & 28.6 & 39.0 & 84.0 & 61.3 & 61.9 & 61.2 \\
& ViT-L/14 & 68.8 & 74.5 & 64.1 & 81.0 & 79.0 & 70.9 & 59.8 & 34.2 & 41.0 & 85.2 & 62.5 & 65.6 & 65.5 \\
\midrule
\multirow{2}{*}{SigLIP} & ViT-B/16 & 44.0 & 67.7 & 58.0 & 70.4 & 64.7 & 64.0 & 63.6 & 25.9 & 37.3 & 79.8 & 56.4 & 55.6 & 57.3 \\
& ViT-L/16 & 55.0 & 71.9 & 63.3 & 76.8 & 71.2 & 68.9 & \underline{73.0} & 26.0 & 40.5 & 84.4 & 58.8 & 58.2 & 62.3 \\
\midrule
\multirow{2}{*}{SigLIP2} & ViT-B/16 & 46.2 & 72.2 & 62.6 & 78.0 & 69.1 & 68.4 & 66.0 & 28.3 & 39.1 & 83.6 & 56.7 & 57.1 & 60.6 \\
& ViT-L/16 & 59.5 & 74.9 & 67.3 & 81.4 & 77.0 & 73.0 & 71.9 & 32.3 & 46.7 & 89.6 & 58.6 & 62.4 & 66.2 \\
\midrule
\multirow{3}{*}{RemoteCLIP} & RN-50 & 21.9 & \underline{83.3} & 47.0 & 70.3 & 49.0 & 60.1 & 65.7 & 22.2 & 33.5 & 93.7 & 59.5 & 66.4 & 56.0 \\
& ViT-B/32 & 42.3 & \textbf{88.9} & 59.1 & 80.8 & 62.0 & 72.3 & 67.0 & 27.1 & 43.0 & \textbf{96.5} & \underline{71.1} & 67.3 & 64.8 \\
& ViT-L/14 & 44.7 & 81.9 & 62.1 & 81.6 & 64.1 & \underline{74.0} & 67.1 & 32.5 & 46.3 & 91.8 & \textbf{71.2} & 63.0 & 65.0 \\
\midrule
\multirow{2}{*}{GeoRSCLIP} & ViT-B/32 & 51.8 & 72.6 & 66.7 & 80.5 & 77.3 & 70.2 & 72.0 & 29.3 & 46.4 & 91.9 & 62.9 & \textbf{75.2} & 66.4 \\
& ViT-L/14 & \underline{70.7} & 77.2 & \textbf{71.7} & \textbf{86.0} & \textbf{81.5} & \textbf{75.1} & \textbf{74.6} & \underline{39.2} & \underline{51.6} & \underline{94.1} & 65.1 & 72.4 & \textbf{71.6} \\
\midrule
CLIP LAION-RS & ViT-L/14 & 70.6 & 73.6 & 65.0 & 79.9 & 75.2 & 69.8 & 65.7 & 35.0 & 46.0 & 90.9 & 63.3 & 68.4 & 67.0 \\
\midrule
SkyCLIP & ViT-L/14 & \textbf{72.2} & 74.3 & \underline{69.1} & \underline{84.2} & \underline{81.2} & 70.6 & 72.9 & {\bfseries 44.9} & {\bfseries 52.8} & 90.0 & 64.4 & \underline{72.6} & \underline{70.8} \\
\bottomrule
\end{tabular}
}
\vspace{-5pt}
\end{table*}
\begin{table*}[ht!]
\centering
\caption{Zero-shot classification accuracy (\%) on all evaluated remote sensing datasets under the meta-prompting approach with \rsllava. Results are reported across the full set of vision-language models and backbone architectures.}
\label{tab:cls_rsllava}
\vspace{-5pt}
\resizebox{\textwidth}{!}{
\begin{tabular}{lcccccccccccccc}
\toprule
\SC{Model} & \SC{Arch}
& \rotatebox{90}{\eurosat}
& \rotatebox{90}{\aid}
& \rotatebox{90}{\mlrsnet}
& \rotatebox{90}{\optimal}
& \rotatebox{90}{\patternnet}
& \rotatebox{90}{\resisc}
& \rotatebox{90}{\rsc}
& \rotatebox{90}{\rsicblow}
& \rotatebox{90}{\rsicbhigh}
& \rotatebox{90}{\whurs}
& \rotatebox{90}{\siriwhu}
& \rotatebox{90}{\rsscn}
& \rotatebox{90}{\SC{Avg}} \\
\midrule
\multirow{2}{*}{CLIP} & ViT-B/16 & 50.4 & 68.3 & 57.0 & 73.6 & 64.4 & 63.8 & 58.2 & 25.5 & 36.4 & 85.1 & 52.9 & 65.9 & 58.5 \\
& ViT-L/14 & 66.7 & 71.9 & 64.8 & 78.7 & 73.8 & 68.7 & 59.8 & 31.9 & 40.7 & 88.5 & 57.9 & 65.8 & 64.1 \\
\midrule
\multirow{2}{*}{MetaCLIP-400M} & ViT-B/16 & 55.1 & 68.4 & 58.6 & 77.1 & 68.7 & 66.3 & 67.3 & 27.6 & 37.3 & 85.6 & 57.0 & 70.0 & 61.6 \\
& ViT-L/14 & 49.9 & 70.6 & 62.7 & 76.8 & 75.3 & 71.4 & 62.3 & 30.0 & 39.1 & 88.4 & 62.6 & 71.4 & 63.4 \\
\midrule
\multirow{2}{*}{MetaCLIP-2.5B} & ViT-B/16 & 50.1 & 71.0 & 63.5 & 78.9 & 72.6 & 68.9 & 62.2 & 31.5 & 38.6 & 85.3 & 60.5 & 70.8 & 62.8 \\
& ViT-L/14 & 61.8 & 74.4 & 68.0 & 82.2 & 77.9 & 72.8 & 60.5 & 37.0 & 39.3 & 88.8 & 62.5 & 70.5 & 66.3 \\
\midrule
\multirow{2}{*}{SigLIP} & ViT-B/16 & 44.4 & 70.7 & 59.1 & 76.5 & 65.1 & 66.0 & 60.9 & 25.8 & 39.1 & 83.3 & 57.9 & 62.3 & 59.3 \\
& ViT-L/16 & 53.9 & 74.7 & 64.3 & 80.9 & 70.9 & 72.4 & \underline{68.4} & 29.3 & 46.1 & 88.1 & 58.7 & 67.4 & 64.6 \\
\midrule
\multirow{2}{*}{SigLIP2} & ViT-B/16 & 46.8 & 73.7 & 63.4 & 82.0 & 69.4 & 71.1 & 62.0 & 29.9 & 42.6 & 87.8 & 59.9 & 65.0 & 62.8 \\
& ViT-L/16 & 60.5 & 76.0 & 69.9 & 83.0 & 75.8 & \underline{75.7} & 66.8 & 33.4 & 46.4 & 91.8 & 60.8 & 69.9 & 67.5 \\
\midrule
\multirow{3}{*}{RemoteCLIP} & RN-50 & 21.7 & 80.8 & 44.2 & 68.5 & 42.0 & 58.6 & 60.4 & 20.1 & 33.4 & 92.8 & 59.4 & 61.4 & 53.6 \\
& ViT-B/32 & 41.1 & \textbf{88.7} & 58.2 & 79.6 & 58.8 & 71.5 & 65.7 & 27.6 & 42.9 & \textbf{95.8} & \underline{71.0} & 64.0 & 63.7 \\
& ViT-L/14 & 45.0 & \underline{81.5} & 60.0 & 80.9 & 62.3 & 73.4 & 65.4 & 29.7 & 47.3 & 90.8 & \textbf{71.3} & 62.2 & 64.2 \\
\midrule
\multirow{2}{*}{GeoRSCLIP} & ViT-B/32 & 52.9 & 73.0 & 65.9 & 81.7 & 76.9 & 71.0 & 66.2 & 32.3 & 46.1 & 91.2 & 61.1 & \textbf{77.5} & 66.3 \\
& ViT-L/14 & 73.3 & 76.8 & \textbf{71.4} & \textbf{86.4} & \underline{79.8} & \textbf{76.6} & \textbf{70.3} & \underline{41.3} & \textbf{55.6} & \underline{93.3} & 61.2 & 72.0 & \textbf{71.5} \\
\midrule
CLIP LAION-RS & ViT-L/14 & \underline{74.3} & 73.9 & 65.0 & 78.6 & 73.8 & 70.2 & 62.5 & 34.8 & 47.3 & 91.7 & 60.6 & \underline{72.6} & 67.1 \\
\midrule
SkyCLIP & ViT-L/14 & \textbf{75.9} & 75.8 & \underline{70.3} & \underline{83.8} & \textbf{81.4} & 73.0 & 68.0 & \textbf{44.7} & \underline{50.1} & 92.5 & 61.6 & 71.4 & \underline{70.7} \\
\bottomrule
\end{tabular}
}
\vspace{-5pt}
\end{table*}
Tables \ref{tab:cls_st}--\ref{tab:cls_rsllava} present the full zero-shot classification results for the additional description-generation strategies, including the Simple Template (\st) baseline and the remaining LLM-based prompt sources DeepSeek, Qwen-2.5, Mixtral, and RS-LLaVA. These tables extend the analysis of Section \ref{sub:classficiation}, where the main paper emphasizes DST and LLaMA-3.1 as the most representative template-based and LLM-based configurations, respectively. Consistent with the main results, the extended tables show that remote-sensing-adapted vision-language models remain the most robust backbones across datasets, while general-purpose models benefit less consistently from prompt variation. The ST baseline is generally weaker than DST, which further supports the importance of domain-aware prompt design even in the absence of meta-prompting. In addition, the remaining LLM-based strategies exhibit the same non-uniform behavior discussed in the main paper: although they can outperform template-based prompts on specific dataset-model pairs, their gains are not systematic and their performance varies substantially across benchmarks. This again highlights that, in the vanilla embedding space, richer LLM-generated descriptions may introduce useful semantic detail but also additional variability that is not always aligned with the discriminative structure of the downstream classes. Overall, the extended results reinforce the main conclusion that prompt effectiveness in zero-shot remote sensing depends on a non-trivial interaction between description source, dataset characteristics, and the underlying VLM.

\clearpage

% --------------------------------------------------
% References
% --------------------------------------------------

%\bibliographystyle{unsrt}
%\bibliography{bib}

\begin{thebibliography}{10}

\bibitem{clip}
Alec Radford, Jong~Wook Kim, Chris Hallacy, Aditya Ramesh, Gabriel Goh,
  Sandhini Agarwal, Girish Sastry, Amanda Askell, Pamela Mishkin, Jack Clark,
  Gretchen Krueger, and Ilya Sutskever.
\newblock Learning transferable visual models from natural language
  supervision.
\newblock In {\em Int. Conf. Mach. Learn.}, volume 139, pages 8748--8763, 2021.

\bibitem{mirza2024mpvr}
M.~Jehanzeb Mirza, Leonid Karlinsky, Wei Lin, Sivan Doveh, , Jakub Micorek,
  Mateusz Kozinski, Hilde Kuhene, and Horst Possegger.
\newblock {Meta-Prompting for Automating Zero-shot Visual Recognition with
  LLMs}.
\newblock In {\em Eur. Conf. Comput. Vis.}, pages 370--387, 2024.

\bibitem{metaprompting2026}
Antonis Promponas, Eirini Baltzi, Valsamis Ntouskos, and Konstantinos
  Karantzalos.
\newblock Meta-prompting with open-source language models for zero-shot scene
  classification in remote sensing.
\newblock In {\em Int. Arch. ISPRS}, page to appear, 2026.

\bibitem{xu2023demystifying}
Hu~Xu, Saining Xie, Xiaoqing~Ellen Tan, Po-Yao Huang, Russell Howes, Vasu
  Sharma, Shang-Wen Li, Gargi Ghosh, Luke Zettlemoyer, and Christoph
  Feichtenhofer.
\newblock Demystifying clip data.
\newblock In {\em Int. Conf. Learn. Represent.}, pages 47812--47831, 2024.

\bibitem{siglip}
Xiaohua Zhai, Basil Mustafa, Alexander Kolesnikov, and Lucas Beyer.
\newblock Sigmoid loss for language image pre-training.
\newblock In {\em Int. Conf. Comput. Vis.}, pages 11975--11986, 2023.

\bibitem{zhou2022learning}
Kaiyang Zhou, Jingkang Yang, Chen~Change Loy, and Ziwei Liu.
\newblock Learning to prompt for vision-language models.
\newblock {\em Int. J. Comput. Vis.}, 130(9):2337--2348, 2022.

\bibitem{zhou2022conditional}
Kaiyang Zhou, Jingkang Yang, Chen~Change Loy, and Ziwei Liu.
\newblock Conditional prompt learning for vision-language models.
\newblock In {\em IEEE Conf. Comput. Vis. Pattern Recog.}, pages 16816--16825,
  2022.

\bibitem{wortsman2022robust}
Mitchell Wortsman, Gabriel Ilharco, Jong~Wook Kim, Mike Li, Simon Kornblith,
  Rebecca Roelofs, Raphael~Gontijo Lopes, Hannaneh Hajishirzi, Ali Farhadi,
  Hongseok Namkoong, et~al.
\newblock Robust fine-tuning of zero-shot models.
\newblock In {\em IEEE Conf. Comput. Vis. Pattern Recog.}, pages 7959--7971,
  2022.

\bibitem{li2018deep}
Ying Li, Haokui Zhang, Xizhe Xue, Yenan Jiang, and Qiang Shen.
\newblock Deep learning for remote sensing image classification: A survey.
\newblock {\em WIRES Data Min. Knowl.}, 8(6):e1264, 2018.

\bibitem{liu2023remoteclip}
Chunhui Liu et~al.
\newblock {RemoteCLIP}: A vision language foundation model for remote sensing.
\newblock {\em IEEE Trans. Geosci. Remote Sens.}, 62:1--16, 2024.

\bibitem{zhang2024rs5m}
Zilun Zhang, Tiancheng Zhao, Yulong Guo, and Jianwei Yin.
\newblock {RS5M and GeoRSCLIP}: A large-scale vision-language dataset and a
  large vision-language model for remote sensing.
\newblock {\em IEEE Trans. Geosci. Remote Sens.}, 62:1--23, 2024.

\bibitem{wang2024skyscript}
Zhecheng Wang, Rajanie Prabha, Tianyuan Huang, Jiajun Wu, and Ram Rajagopal.
\newblock Skyscript: A large and semantically diverse vision-language dataset
  for remote sensing.
\newblock {\em AAAI}, 38(6):5805--5813, 2024.

\bibitem{lu2017exploring}
Xiaoqiang Lu, Binqiang Wang, Xiangtao Zheng, and Xuelong Li.
\newblock Exploring models and data for remote sensing image caption
  generation.
\newblock {\em IEEE Trans. Geosci. Remote Sens.}, 56(4):2183--2195, 2017.

\bibitem{llama3_1}
Aaron Grattafiori, Abhimanyu Dubey, Abhinav Jauhri, Abhinav Pandey, Abhishek
  Kadian, Ahmad Al-Dahle, Aiesha Letman, Akhil Mathur, Alan Schelten, Alex
  Vaughan, et~al.
\newblock {The Llama 3 Herd of Models}, 2024.

\bibitem{bi2024deepseek}
DeepSeek-AI, :, Xiao Bi, Deli Chen, Guanting Chen, Shanhuang Chen, Damai Dai,
  Chengqi Deng, Honghui Ding, Kai Dong, Qiushi Du, Zhe Fu, et~al.
\newblock {DeepSeek LLM: Scaling Open-Source Language Models with Longtermism},
  2024.

\bibitem{qwen2_5}
Binyuan Hui, Jian Yang, Zeyu Cui, Jiaxi Yang, Dayiheng Liu, Lei Zhang, Tianyu
  Liu, Jiajun Zhang, Bowen Yu, Keming Lu, Kai Dang, Yang Fan, Yichang Zhang,
  An~Yang, Rui Men, Fei Huang, Bo~Zheng, Yibo Miao, Shanghaoran Quan, Yunlong
  Feng, Xingzhang Ren, Xuancheng Ren, Jingren Zhou, and Junyang Lin.
\newblock Qwen2.5-coder technical report, 2024.

\bibitem{mixtral}
Albert~Q. Jiang, Alexandre Sablayrolles, Antoine Roux, Arthur Mensch, Blanche
  Savary, Chris Bamford, Devendra~Singh Chaplot, Diego de~las Casas, et~al.
\newblock Mixtral of experts, 2024.

\bibitem{bazi2024rs}
Yakoub Bazi, Laila Bashmal, Mohamad~Mahmoud Al~Rahhal, Riccardo Ricci, and
  Farid Melgani.
\newblock Rs-llava: A large vision-language model for joint captioning and
  question answering in remote sensing imagery.
\newblock {\em Remote Sensing}, 16(9):1477, 2024.

\bibitem{pmlr-v267-levi25b}
Meir~Yossef Levi and Guy Gilboa.
\newblock The double-ellipsoid geometry of {CLIP}.
\newblock In {\em Int. Conf. Mach. Learn.}, volume 267, pages 33999--34019,
  2025.

\bibitem{siglip2}
Michael Tschannen, Alexey Gritsenko, Xiao Wang, Muhammad~Ferjad Naeem, Ibrahim
  Alabdulmohsin, Nikhil Parthasarathy, Talfan Evans, Lucas Beyer, Ye~Xia, Basil
  Mustafa, Olivier Hénaff, Jeremiah Harmsen, Andreas Steiner, and Xiaohua
  Zhai.
\newblock Siglip 2: Multilingual vision-language encoders with improved
  semantic understanding, localization, and dense features, 2025.

\bibitem{betser2026whitened}
Roy Betser, Meir~Yossef Levi, and Guy Gilboa.
\newblock Whitened {CLIP} as a likelihood surrogate of images and captions.
\newblock In {\em Int. Conf. Mach. Learn.}, volume 267, pages 4069--4095, 2025.

\bibitem{zou2015deep}
Qin Zou, Lihao Ni, Tong Zhang, and Qian Wang.
\newblock Deep learning based feature selection for remote sensing scene
  classification.
\newblock {\em {IEEE Geoscience and Remote Sensing Letters}},
  12(11):2321--2325, 2015.

\bibitem{helber2019eurosat}
Patrick Helber, Benjamin Bischke, Andreas Dengel, and Damian Borth.
\newblock Eurosat: A novel dataset and deep learning benchmark for land use and
  land cover classification.
\newblock {\em IEEE J. Sel. Top. Appl. Earth Obs. Remote Sens.},
  12(7):2217--2226, 2019.

\bibitem{zhao2016feature}
Lijun Zhao, Ping Tang, and Lianzhi Huo.
\newblock Feature significance-based multibag-of-visual-words model for remote
  sensing image scene classification.
\newblock {\em J. Appl. Remote Sens.}, 10(3):035004--035004, 2016.

\bibitem{zhao2015dirichlet}
Bei Zhao, Yanfei Zhong, Gui-Song Xia, and Liangpei Zhang.
\newblock Dirichlet-derived multiple topic scene classification model for high
  spatial resolution remote sensing imagery.
\newblock {\em IEEE Trans. Geosci. Remote Sens.}, 54(4):2108--2123, 2015.

\bibitem{xia2010structural}
Gui-Song Xia, Wen Yang, Julie Delon, Yann Gousseau, Hong Sun, and Henri
  Ma{\^\i}tre.
\newblock Structural high-resolution satellite image indexing.
\newblock In {\em ISPRS TC VII Symposium}, volume~38, pages 298--303, 2010.

\bibitem{xia2017aid}
Gui-Song Xia, Jingwen Hu, Fan Hu, Baoguang Shi, Xiang Bai, Yanfei Zhong,
  Liangpei Zhang, and Xiaoqiang Lu.
\newblock Aid: A benchmark data set for performance evaluation of aerial scene
  classification.
\newblock {\em IEEE Trans. Geosci. Remote Sens.}, 55(7):3965--3981, 2017.

\bibitem{wang2018scene}
Qi~Wang, Shaoteng Liu, Jocelyn Chanussot, and Xuelong Li.
\newblock Scene classification with recurrent attention of vhr remote sensing
  images.
\newblock {\em IEEE Trans. Geosci. Remote Sens.}, 57(2):1155--1167, 2018.

\bibitem{li2020rsi}
Haifeng Li, Xin Dou, Chao Tao, Zhixiang Wu, Jie Chen, Jian Peng, Min Deng, and
  Ling Zhao.
\newblock {RSI-CB: A large-scale remote sensing image classification benchmark
  using crowdsourced data}.
\newblock {\em Sensors}, 20(6):1594, 2020.

\bibitem{zhou2018patternnet}
Weixun Zhou, Shawn Newsam, Congmin Li, and Zhenfeng Shao.
\newblock Patternnet: A benchmark dataset for performance evaluation of remote
  sensing image retrieval.
\newblock {\em ISPRS J. Photogramm. Remote Sens.}, 145:197--209, 2018.

\bibitem{cheng2017remote}
Gong Cheng, Junwei Han, and Xiaoqiang Lu.
\newblock Remote sensing image scene classification: Benchmark and state of the
  art.
\newblock {\em Proceedings of the IEEE}, 105(10):1865--1883, 2017.

\bibitem{qi2020mlrsnet}
Xiaoman Qi, Panpan Zhu, Yuebin Wang, Liqiang Zhang, Junhuan Peng, Mengfan Wu,
  Jialong Chen, Xudong Zhao, Ning Zang, and P~Takis Mathiopoulos.
\newblock Mlrsnet: A multi-label high spatial resolution remote sensing dataset
  for semantic scene understanding.
\newblock {\em ISPRS J. Photogramm. Remote Sens.}, 169:337--350, 2020.

\end{thebibliography}

\end{document}